\documentclass[preprint,12pt,authoryear,nonatbib]{elsarticle}

\makeatletter
\let\c@author\relax
\makeatother

\usepackage{amssymb}
\usepackage{amsmath}
\usepackage{xcolor}

\usepackage[acronym]{glossaries}
\newacronym{bnf}{BNF}{element-wise Bernstein normalizing flow}
\newacronym{ctm}{CTM}{conditional transformation model}
\newacronym{cdf}{CDF}{cummulative density function}
\newacronym{nf}{NF}{normalizing flow}
\newacronym{rnn}{RNN}{recurrent neural network}
\newacronym{pit}{PIT}{probability integral transform}
\newacronym{nll}{NLL}{negative logarithmic likelihood}
\newacronym{WGAN}{WGAN}{hybrid Wasserstein Generative Adversarial Network}
\newacronym{DDPM}{DDPM}{Denoising Diffusion Probabilistic Model}
\newacronym{MABF}{MABF}{Masked Autoregressive Bernstein polynomial normalizing Flows}
\newacronym{MADE}{MADE}{masked autoencoder for distribution estimation}
\newacronym{MAF}{MAF}{masked autoregressive flow}
\newacronym{HMM}{HMM}{Hidden Markov Model}
\newacronym{SLP}{SLP}{synPro Standard Load Profile}
\newacronym{umap}{UMAP}{Uniform Manifold Approximation and Projection}
\newacronym{om}{OM}{OpenMeter}

\usepackage{bm}

\usepackage[english,ruled,vlined]{algorithm2e}
\usepackage{algorithmic}
\usepackage{subcaption}

\journal{}

\begin{document}

\begin{frontmatter}



\title{Time-series surrogates from energy consumers generated by machine learning approaches for long-term forecasting scenarios}

\author[a,b]{Ben Gerhards}
\author[a]{Nikita Popkov}
\author[c]{Annekatrin König}
\author[d,c]{Marcel Arpogaus}
\author[b]{Bastian Schäfermeier}
\author[b]{Leonie Riedl}
\author[a]{Stephan Vogt}
\author[c]{Philip Hehlert\corref{cor1}}

\cortext[cor1]{Corresponding author: philip.hehlert@uni-goettingen.de}

\affiliation[a]{organization={Intelligente Eingebettete Systeme Universität Kassel},
            addressline={Wilhelmshöher Allee 73}, 
            city={Kassel},
            postcode={34121}, 
            state={Hesse},
            country={Germany}}
            
\affiliation[b]{organization={Fraunhofer IEE},
            addressline={Joseph-Beuys-Straße 8}, 
            city={Kassel},
            postcode={34117}, 
            state={Hesse},
            country={Germany}}

\affiliation[c]{organization={Georg-August Universität Göttingen},
            addressline={Julia-Lermontowa-Weg 3}, 
            city={Göttingen},
            postcode={37077}, 
            state={Lower Saxony},
            country={Germany}}

\affiliation[d]{organization={HTGW Hochschule Konstanz},
            addressline={Alfred-Wachtel-Str. 8}, 
            city={Konstanz},
            postcode={78462}, 
            state={Baden-Württemberg},
            country={Germany}}

\begin{abstract}
Forecasting attracts a lot of research attention in the electricity value chain. However, most studies concentrate on short-term forecasting of generation or consumption with a focus on systems and less on individual consumers. Even more neglected is the topic of long-term forecasting of individual power consumption.

Here, we provide an in-depth comparative evaluation of data-driven methods for generating synthetic time series data tailored to energy consumption long-term forecasting. High-fidelity synthetic data is crucial for a wide range of applications, including state estimations in energy systems or power grid planning. In this study, we assess and compare the performance of multiple state-of-the-art but less common techniques: a \gls{WGAN}, \gls{DDPM}, \gls{HMM}, and \gls{MABF}. \\ 
We analyze the ability of each method to replicate the temporal dynamics, long-range dependencies, and probabilistic transitions characteristic of individual energy consumption profiles. Our comparative evaluation highlights the strengths and limitations of: \gls{WGAN}, \gls{DDPM}, \gls{HMM} and \gls{MABF} aiding in selecting the most suitable approach for state estimations and other energy-related tasks. Our generation and analysis framework aims to enhance the accuracy and reliability of synthetic power consumption data while generating data that fulfills criteria like anonymisation - preserving privacy concerns mitigating risks of specific profiling of single customers. This study utilizes an open-source dataset from households in Germany with 15min time resolution. The generated synthetic power profiles can readily be used in applications like state estimations or consumption forecasting.
\end{abstract}




\end{frontmatter}




\section{Intro}
The electricity sector is currently transitioning from the traditional scheme of a central producer and consumers into a complex system of decentralized generators, energy stores, and dynamic prosumers. This shift creates new challenges for future electricity grids. Thus, for reliable planning and control, forecasting and particularly long-term forecasting of energy loads becomes imperative. Whereas various short-term prediction approaches exist (reviewed by \cite{review_hong_2020,review_scheidt_2020}), few studies focus on long-term load forecasting \cite{kim_cho_2019,wang_lstm_2020,salinas_deepar_2020,herrero_2022}. This might be caused by the challenging nature of time series data with cyclic patterns on a daily basis (micro-scale) and long-term effects (macro-scale) like seasonal variations on electricity loads or the installation of solar-panels. All these effects are combined with fluctuations in daily behavior and multiple external influencing factors like the weather. In recent years, various machine learning approaches were established to model individual load profiles.\\

Creating realistic synthetic load profiles requires accurate data. However, despite the ongoing modernization from classical electromechanical meters to smart meters, data sources are scarce and laden with concerns on privacy. Additionally, the global smart meter adoption that allows for the continuous measuring of power consumption is still in its early phase \cite{aryandoust_2022}. For example, despite a yearly investment of $\sim$750 Mio Euro (for 2023) into power meters \cite{monitor_germany_2023} (around 50\% on smart meters) roughly 0.5\% of power connections in Germany have intelligent smart meters that allow for continuous real-time monitoring. 
Additionally, legislation like General Data Protection Regulation (GDPR)\cite{GDPR_2016} compliance produce administrative barriers that hinder an easy-accessible data aggregation for research purposes. In combination these effects result in a strong velocity constraint for the data \cite{aryandoust_2022,kezunovic_2013,yu_2015}. However, there is a growing community and platforms like OpenMeter \cite{albrecht_openmeter_2024} are emerging that provide open access to load profiles.\\

Digital surrogates, synthetic load profiles that preserve characteristics of individual consumers, could overcome some of the issues. In this study we introduce a complete framework of tools (written in Python) to classify consumer types, analyze load-profile characteristics and train a selection of machine learning models to generate and validate synthetic load profiles in a long-term forecasting scenario and compare it to generally utilized standard load profiles \cite{meier_reprasentative_1999}.

\section{Dataset, consumer types and metrics}
\subsection{OpenMeter dataset}

The OpenMeter Data Platform was created to address the growing need for publicly accessible, real-world energy consumption data and represents the first large, open access, dataset of energy prosumers from Germany (ongoing data collection). The load profiles cover a time span from 0.5 to up to 20 years (median: 3.4 years, by January 2025). Energy consumption data and meta data are contributed by both public institutions and private entities, including households, businesses, and industrial facilities. As prosumers, database contributors can also provide discrete photovoltaic production data.

The consumption profiles include 4 measurements per hour (15min resolution), and the available metadata include address (postal code area and district), the date, and the consumer category (e.g. private household, public building, commercial). Hitherto, the database includes approximately 1,000 private households, about 200 public institutions and about 60 businesses that are distributed all over Germany. SI units, such as kilowatts (kW) rather than watts (W) are used for data storage. UTC, or Coordinated Universal Time, is used since it does not include night or day shifts or any adjustments for daylight saving time.
For optimal machine learning processes, the data had to be cleaned and filtered. Only load data that were measured over a period of at least 2.5 years and 15min resolution were included. The number of missing values (NaN: "Not a number") was quantified and only datasets with a maximum of 5\% NaN values were used. In addition, implausibly high values -- such as those caused by faulty conversion of meter readings to electrical consumption -- were removed, along with multiple consecutive zeros. Geographical information was verified for plausibility, and leading zeros in postal codes were corrected. Data normalization was performed via min-max scaling, except when individual scaling was required for model training. Finally, the OM dataset was divided into a 1.5-year training set and a 1-year test set.

\subsection{Consumer typing}
While individual load profiles differ, common characteristics and patterns typically emerge in large populations. Clustering consumers by their energy usage patterns allows for the generation of labels and the formation of groups. This can be used to optimize energy distribution, grid planning, or identifying inefficiencies of outliers or groups. Furthermore, labels can be used to segment different consumers for e.g. hierarchical machine learning or serve as an input for consumer-type specific inference. Commonly, static metrics are isolated from individual consumers and subsequently clustered \cite{benitez2017classification,rasanen2009feature,yilmaz2019comparison,shi2020approach,silipo2013big}. A recent approach utilized a principal component analysis (PCA) on typical weeks and exhibited robust group segmentation with the use of principal component one to five (PC1-5) \cite{riedl2024cigre}. 

\subsection{Metrics}
Hitherto, no general benchmark exists to evaluate the quality of long-term synthetic load profiles \cite{review_scheidt_2020,hong_review_2020}. For comparability and best practice we will employ various metrics: Point-features like root mean squared error (RMSE), mean absolute percentage error (MAPE) or mean absolute squared error (MASE)  will be analyzed. In addition, with the focus on realistic synthetic load profiles we are also interested in the distributions of the loads and will extract classic parameters like: mean, median, min-, max-values and employ the maximum mean discrepancy (MMD) \cite{gretton2006kernel}. Statistical features used for the consumer typing will be utilized in the calculation of correlation coefficients to check if basic characteristics of individual consumers remain in the synthetic load profiles.
Additionally the synthetic data will be classified by the original load profiles (OM data) labels and a UMAP analysis \cite{mcinnes1802umap} will be applied on daily samples to visualize clusters and structures within the data and evaluate the overlap between the OM (original) and synthetic load profiles.

\section{Models}
Time series forecasting is a highly active research area with numerous approaches being developed across various domains such as finance, healthcare, climate science or the energy sector. While energy consumption forecasting is a well-explored area (review \cite{hong_review_2020, review_scheidt_2020, nti_electricity_2020}), much of the focus has been on short-term forecasting. In contrast, long-term forecasting, particularly at the individual consumer level, has received significantly less attention due to its inherent complexity, evolving consumption patterns and external uncertainties.

Early forecasting methods predominantly relied on autoregressive models such as ARIMA \cite{box_2016}, which remains a widely used baseline model \cite{yunfan_li_2024, de_oliveira_arima_2018}. The introduction of machine learning techniques, including Support Vector Machines~\cite{dong_2005}, Random Forests~\cite{jiao_2022}, and Decision Trees~\cite{tso_yan_2007}, enabled the capture of more complex consumption patterns, though these methods remained largely confined to short-term forecasting. With the rise of deep learning, more sophisticated models have gained prominence, expanding forecasting capabilities. These include Feed-forward Neural Networks (NNs) \cite{tso_yan_2007}, Autoencoders~\cite{li_building_2017}, Fuzzy Logic Systems~\cite{souhe_2021}, and Bayesian Networks~\cite{singh_2018, jiao_2022}.
\\
More powerful architectures, such as Convolutional Neural Networks (CNNs) \cite{muralitharan_2018} and Recurrent Neural Networks (RNNs) \cite{rahman_2018, salinas_deepar_2020}, have demonstrated greater stability and scalability when applied to large datasets. However, the most flexible and widely adopted models for mid-term forecasting include Transformer networks~\cite{vaswani_attention_2017,li_transformer_2019} and RNN variants, particularly Long Short-Term Memory (LSTM) networks~\cite{wang_lstm_2020,qureshi_LSTM_2024}. Notably, the development of hybrid models, such as CNN-RNN architectures~\cite{lai_liu_2018, kim_cho_2019,aryandoust_2022,herrero_2022,ramos_2023,yunfan_li_2024, ou_ali_autoencoder_2024} and Transformer-based variants~\cite{wang_lstm_informer_2023, wu_autoformer_2021,zhou_informer_2021}, has extended forecasting horizons. However, these methods remain limited to projections of less than a year and thus usually have limited resolution.

In this work, we explore machine learning approaches with a particular emphasis on recent advancements in generative AI. These methods offer promising capabilities for long-term forecasting, enabling probabilistic load estimation rather than simple point forecasts and allowing for sampling from the underlying load distribution.

This makes them particularly valuable in applications that require multi-scale insights, ranging from individual consumer demand to aggregated power loads. Furthermore, the models we compare leverage meta-data and external influencing factors like weather conditions. By incorporating meta-data, such as postal code and consumer type as input features, synthetic power profiles can be generated for configurable scenarios, enhancing their applicability in energy system planning and simulation.

\subsection{Hybrid Wasserstein Generative Adversarial Network (WGAN)}
For the generation of synthetic power profiles we chose an advanced generative model that was tailored to synthesize realistic time-series data. Our approach is based on a Wasserstein GAN, that addresses instability and mode collapses often encountered in traditional GANs by employing the Wasserstein distance as the optimization metric \cite{arjovsky2017wasserstein}. Additionally gradient penalty is employed to further stabilize the training \cite{gulrajani2017improved}.

The generator is implemented by utilizing a Long Short-Term Memory (LSTM) network \cite{hochreiter1997long} to capture temporal dependencies inherent in time-series data. The discriminator is designed as a feed-forward neural network. To refine the generative process both, generator and discriminator, are receiving input from conditional blocks, that process meta-data like seasonal or temporal features but can also have geographical information, weather data or consumer type (Figure \ref{fig:CWGAN}). 
With this architecture the \gls{WGAN} combines the strengths of adversarial learning and sequence modeling and should allow the model to capture statistical and temporal properties of the real-world consumers but also possess sufficient variability to ensure anonymization.

\subsection{Denoising Diffusion Probabilistic Models (DDPMs)}

\begin{figure}[!ht]
\centering
\resizebox{\textwidth}{!}{\includegraphics{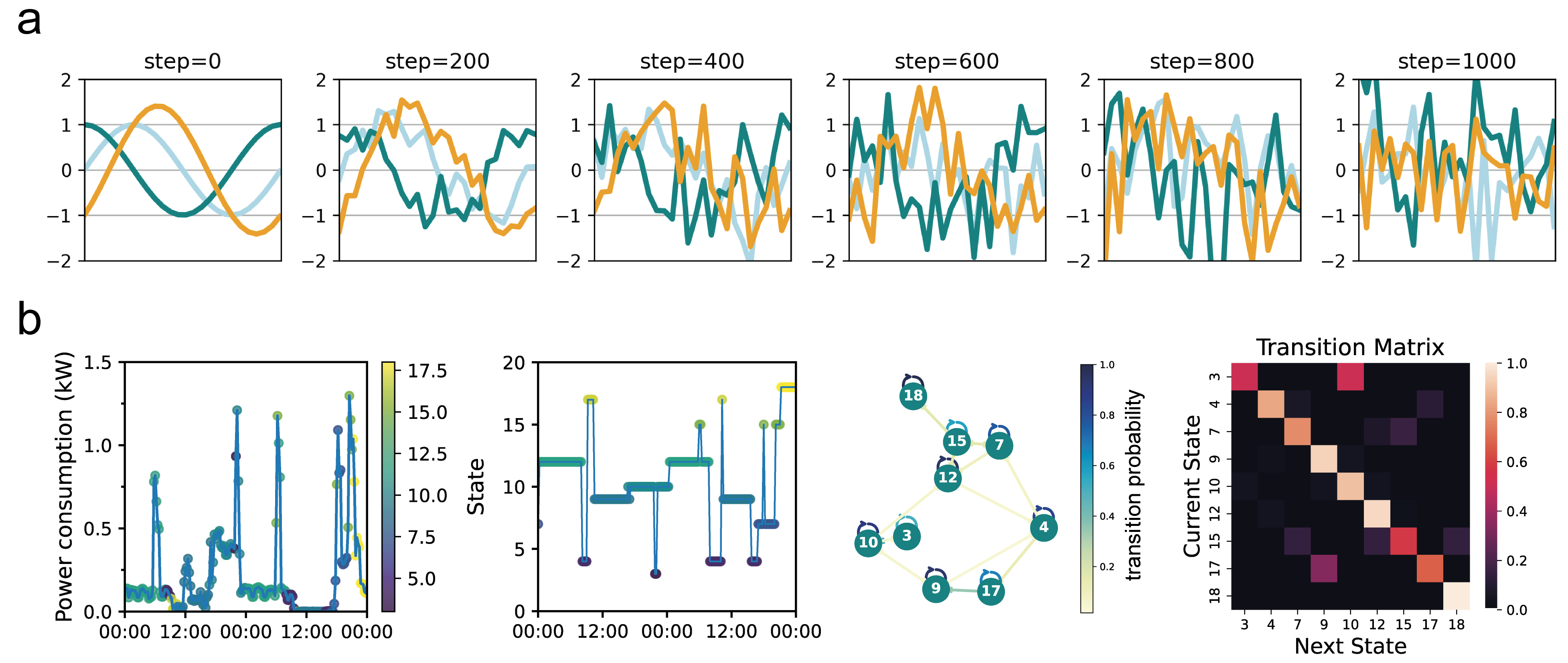}}
\caption{DDPM and HMM principles - a) Gradually adding noise to a sample as an example for the DDPM forward pass. b) Self-classification of continuous power consumption data into sequence of states. Transition probabilities can be calculated from the self-classification (Network graph; Transition Matrix) to build a Hidden Markov Model (HMM).}\label{fig:noising example}
\end{figure}

Denoising Diffusion Probabilistic Models (DDPMs) \cite{HoJainAbbeel2020, NicholDhariwall2021} are a class of generative models originally developed for high-dimensional image synthesis. Unlike GANs, which generate samples in a single forward pass, DDPMs rely on a forward–reverse diffusion mechanism: Gaussian noise is gradually added to real data over multiple steps (forward process), and a learned reverse process incrementally de-noises this signal to reconstruct samples. While this concept was initially applied to 2D imagery, its underlying probabilistic formulation and stochastic sampling framework are particularly attractive for modeling continuous, high-resolution time-series data, where capturing temporal dependencies and realistic variation is essential.

The use of DDPMs for time-series generation is relatively novel and not widely explored. Compared to autoregressive and transformer-based models, which often require explicit modeling of sequence dependencies and may suffer from exposure bias or overfitting in long-range settings, DDPMs provide a principled way of modeling the full data distribution without relying on strict temporal factorization. This allows them to generate diverse and coherent samples over extended horizons, an essential requirement for realistic long-term power consumption synthesis. Moreover, DDPMs are inherently probabilistic, offering advantages for uncertainty quantification through their ability to generate novel samples beyond the training data distribution, as well as scenario generation over deterministic or point-forecasting models.

In this work, we adapt the DDPM framework to energy consumption data by modeling the time-series as continuous trajectories and using a 1D U-Net architecture (as an adaptation from 2D-U-Nets in image generation to one-dimensional time series data) as the noise predictor. U-Nets have multiple advantages like skip connections and multi-resolution processing, and hierarchical feature learning. The model is trained to reverse a Markovian degradation process that adds noise to real load profiles. Through this iterative de-noising, we generate realistic surrogates that maintain temporal structure and statistical fidelity. Figure \ref{fig:noising example} (upper row) illustrates the progressive noising process and its impact on signal degradation.

\subsection{Hidden Markov Models (HMMs)}
HMMs have been successfully used in behavior analysis \cite{geurten_hmm_2010,braun_hmm_2010,gabrielski_markov_2023} and provide a powerful forecasting tool due to their ability to model a sequence of unknown, thus hidden, states and observable outcomes. Power loads follow complex patterns that depend on a multitude of external factors (features) like the weather or time of the day. HMMs can model these patterns by assuming that the observed load at a given time is the result of a an underlying hidden state. By quantizing the given load and features at each time-point and subsequent clustering, a process called continuous self-classification, one can identify groups of hidden states. The transition probabilities between these states then enable forecasting (Figure \ref{fig:noising example}, lower row). 

\subsection{MABF}
\glspl{MABF} combine two powerful concepts: transformation models~\cite{Hothorn2014,Hothorn2018} and autoregressive normalizing flows~\cite{Papamakarios2018,Papamakarios2021}.
At its core this model uses flexible Bernstein Polynomials\cite{Farouki2012} to transform data from a complex target distribution into a simple base distribution, like a standard Gaussian.
The transformation is handled in an autoregressive way, where the probability density function of each time-point of the data is conditioned on preceding data points, preserving temporal dependencies.
Specifically the multivariate conditional density $f_{\mathbf{Y}|\mathbf{X}}$ is factorized 
with the chain rule of probability, allowing robust likelihood-based optimisation.

\subsection{SLP}
As a baseline model, for comparison, we used modified VDEW standard load profiles (H0N) \cite{meier_reprasentative_1999} that are typically used by network operators for grid planning, control or power procurement on the energy market. The synPRO generator \cite{fischer_synpro_2020} was employed to generate power consumption data (from here on: SLP) for a whole year for a family of four. Subsequently this profile was min-max-scaled to the median of the 5\% \textit{highest} and \textit{lowest} power consumption of a respective power consumer from the OM dataset.

\section{Results}
\subsection{Visual diagnostics}
\begin{figure}[!ht]
\centering
\resizebox{0.95\textwidth}{!}{\includegraphics{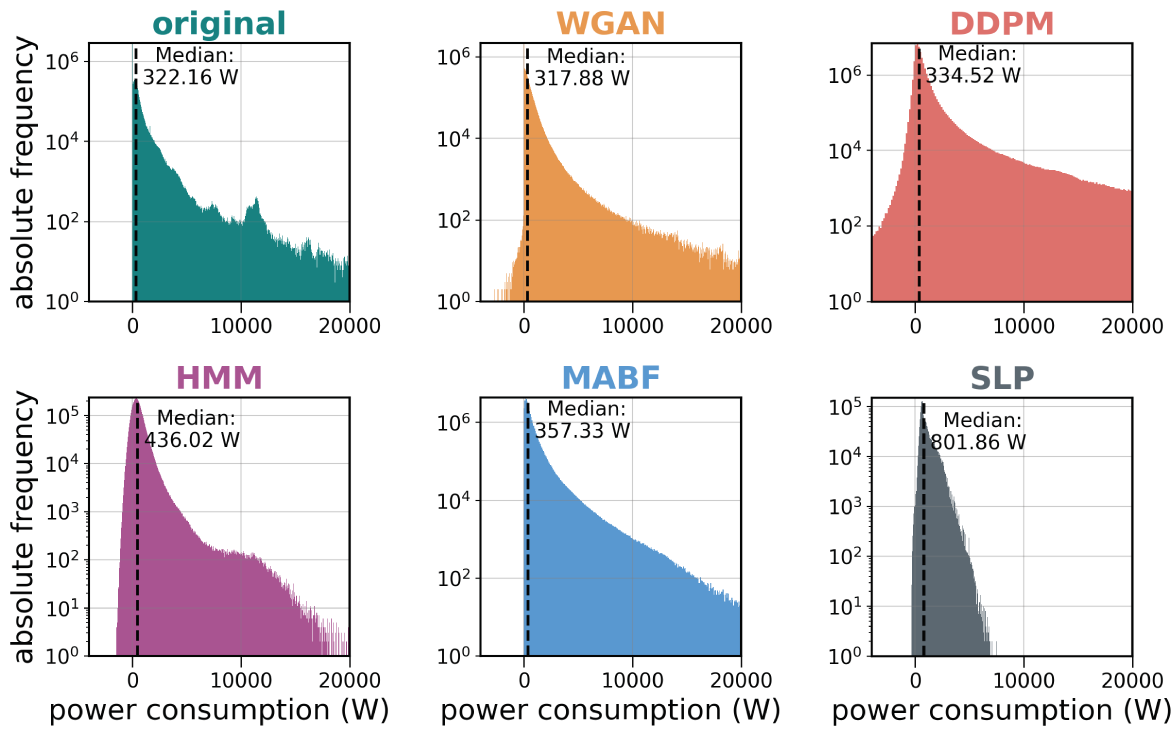}}
\caption{Distribution of power consumption - Absolute frequency of power consumption values from all 552 households of the OM dataset (original) and their respective time-series surrogates (from \gls{WGAN}, \gls{DDPM}, \gls{HMM}, \gls{MABF}, and \gls{SLP}).}
\label{fig:year_distr}
\end{figure}

First, the different distributions of loads were analysed for the different ML models and compared to the ground truth (trained original data from the \gls{om} dataset). For this, frequencies of power consumption were analysed over the test period for each time point (Figure \ref{fig:year_distr}). The power consumers of the OM dataset (original) had a median power consumption of $\sim$320W which was preserved by the \gls{WGAN} and \gls{DDPM} with a deviation of less then 5\%. The \gls{MABF} exhibited a slight shift that resulted in +10\% median power consumption. Power loads of the  \gls{HMM} (+35\%) and the \gls{SLP} (+148\%) were significantly increased (Suppl. Table \ref{tab:simple_stats}, Figure \ref{fig:easy_stats}). All but one model (\gls{MABF}) generated also negative power consumption values that are non-existent in the original data (Suppl. Table \ref{tab:simple_stats}, Figure \ref{fig:easy_stats}).

\begin{figure}[!ht]
\centering
\resizebox{0.99\textwidth}{!}{\includegraphics{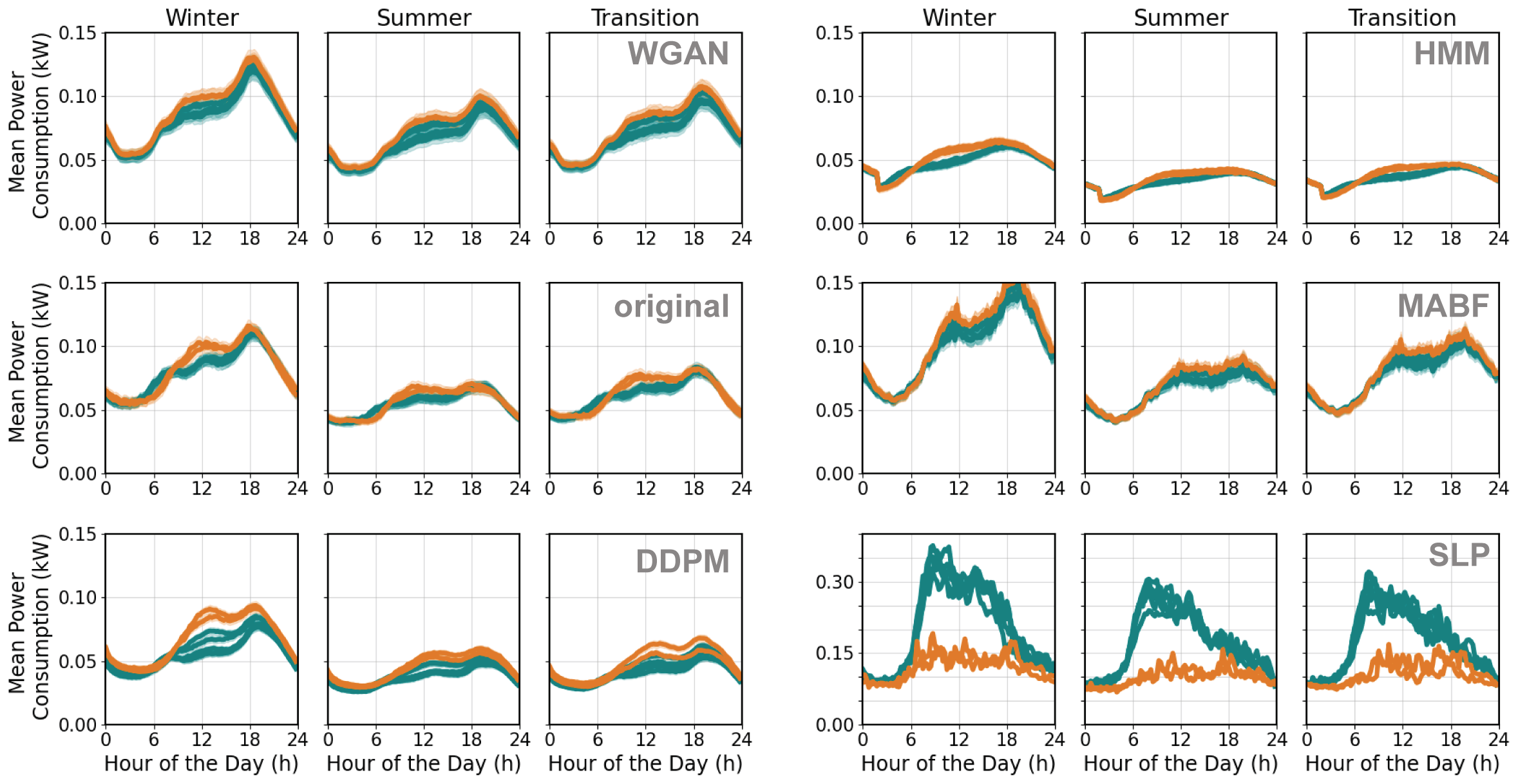}}
\caption{Daily power consumption - Shown are daily averages $\pm95$\% CI (reduced transparency) from typical weeks for the respective ML-models (\gls{WGAN}, \gls{DDPM}, \gls{HMM}, \gls{MABF},\gls{SLP}).(color code: teal= Mon - Fri, orange= Sat, Sun)}
\label{fig:clust_analysis}
\end{figure}

An average electricity consumption (+95\% confidence interval) was then calculated for each day of the week for each point in time of the day from all profiles to get a visual impression of how well the different ML models preserve the daily patterns of electricity consumption. Despite the \gls{SLP} all ML models exhibited a circadian rhythm that was close to the OM (original) data (Figure \ref{fig:clust_analysis}). The characteristic differences between working-days and weekends as observed in the original data were preserved by the ML models. 

Subtracting the OM (original) power consumption values for each time-point of the day from a ML model, allows to analyse at which time of the day the respective ML models generate power over- or under-consumption (Figure \ref{fig:avg_power_diff}). Here, the \gls{WGAN} had the tendency to generate higher values during the evening peak-power consumption and lower consumption in the morning. The \gls{DDPM} overall generated higher power consumption values that were more pronounced in periods of increased power consumption of the original profiles (Figure \ref{fig:avg_power_diff}b).

Quantile-Quantile Plots (Q-Q Plots) are a common visual tool used to further assess how well the original data distribution is matched. In Figure~\ref{fig:qq_plot} the empirical data quantiles are plotted against the empirical quantiles obtained from the generated data.
The figure confirms that WGAN, HMM, and MABF seem to be well-calibrated for low and medium loads, while they appear to overestimate peak loads. Additionally, it becomes evident that only MABF is correctly modeling the truncated strictly positive distribution of the data.

The \gls{HMM} used the average as a starting point and generated deviations around it. Thus, it followed the average of the original profile closely. However, the  \gls{HMM} was not able to generate the same variation in power consumption like the original data. Comparable to the \gls{WGAN}, the \gls{MABF} had the same tendency to generate under-consumption in the morning and over-consumption at periods of increased power consumption of the original data but deviations were more pronounced (Figure \ref{fig:avg_power_diff}b). At last, the utilized \gls{SLP} consumers exhibited a completely different daily power consumption pattern on working day compared to the OM households, but surprisingly was close on weekend days. Thus, \gls{SLP} consumers are not really representative for the analyzed OM households underlining the need of ML models to generate power consumption data that is closer to the ground truth.

\subsection{Seasonality}
Private households in Germany usually have seasonal changes in their energy consumption pattern, with increased power consumption in the winter months and lower consumption during the summer. The data was therefore transformed and analysed as typical weeks from three seasons: winter, summer and transition period (Figure \ref{fig:clust_analysis}). Comparable to the power consumption analysis for each weekday (Figure \ref{fig:avg_power_diff}) the daily pattern of power consumption persists throughout the year but shows different magnitudes for each season with winter being the highest. All ML-models (\gls{WGAN}, \gls{DDPM}, \gls{HMM}, \gls{MABF}) were able to capture this pattern, with the DDPM generating the most accurate time-series surrogates in terms of seasonality (Figure \ref{fig:clust_analysis}).\\

Times of over- or underconsumption as observed earlier by the different models (Figure \ref{fig:avg_power_diff}) could not be attributed to a specific season but persisted in general for synthetic load profiles from the respective models (Figure \ref{fig:clust_analysis}).

In contrast, the \gls{SLP} data did not show pronounced seasonal changes in the power consumption pattern (Figure \ref{fig:clust_analysis}).

\subsection{Dimensionality reduction of daily power consumption patterns}

\begin{figure}[!ht]
\centering
\resizebox{1.0\textwidth}{!}{\includegraphics{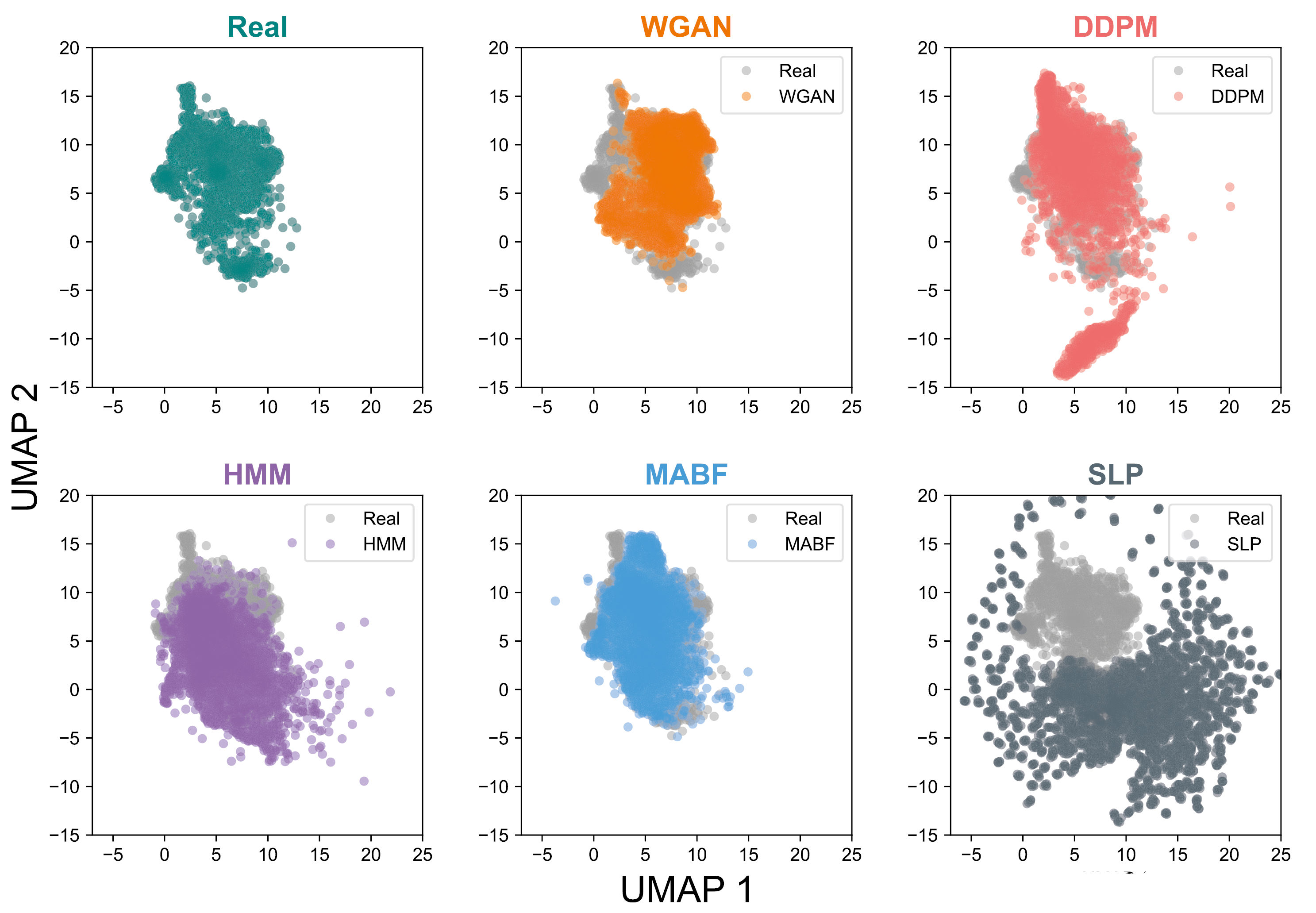}}
\caption{Distribution of daily samples - UMAP analysis of single day samples. 2D-projection of daily power consumption samples from the compared ML-models (WGAN, DDPM, HMM, MABF). With the exception SLP all models were able to capture the distribution of real daily samples.}
\label{fig:umap}
\end{figure}

In order to explore and visualize patterns in power consumption, the time series data was transformed into daily samples. Each sample consisted of power consumption data (15min sampling rate) over 24h. The samples from all models were then analyzed using \gls{umap}, a non-linear dimensionality reduction technique, to reveal underlying patterns and relationships in the data.
The \gls{umap} analysis revealed no obvious distinct clusters within the real consumption data. Most daily samples where quite similar to each other and seasonal effects that can be seen in the averages (Figure \ref{fig:clust_analysis}) appear to have overlapping transition periods and no clearly distinguishable rapid daily changes in energy consumption (Figure \ref{fig:umap}).  Surprisingly, temperature had no strong effect on \gls{umap}. Grouping samples by season neither altered the overall shape nor the density of the projection (Suppl. Figure \ref{fig:umap-temperature}). However, this has been partially expected, as the overall daily consumption pattern of a consumer usually persists throughout all seasons (see Figure \ref{fig:clust_analysis}).

The limited similarity of the \gls{SLP} samples, as observerd earlier, consequently had a strong effect in the UMAP analysis. There was only a very limited overlap of the \gls{SLP} samples with real days from the OM dataset (Figure \ref{fig:umap}). In contrast, all ML models where able to create quite realistic daily power consumption samples and covered most of the days found in the real data. Thus, overall patterns of the underlying daily power consumption patterns where efficiently reproduced by the various ML models (Figure \ref{fig:umap}).

Best coverage could be achieved by the \gls{WGAN} and \gls{MABF} with the lowest average euclidean distance of Hungarian matched daily real and synthetic power-consumption day samples (Suppl. Table \ref{tab:umap_table}). While covering most real daily samples, the \gls{DDPM} and \gls{HMM} both generated synthetic daily power consumption samples that cannot be found in the OM data. The \gls{DDPM} in particular exhibited a smaller median euclidean distance to Hungarian paired daily samples than the \gls{HMM} (Suppl. Table \ref{tab:umap_table}), but created potentially a sub-cluster of day samples that have no similarity to any day found in the OM dataset (Figure \ref{fig:umap}).\\

\subsection{Consumer typing and stability}
\begin{figure}[!ht]
\centering
\resizebox{1.0\textwidth}{!}{\includegraphics{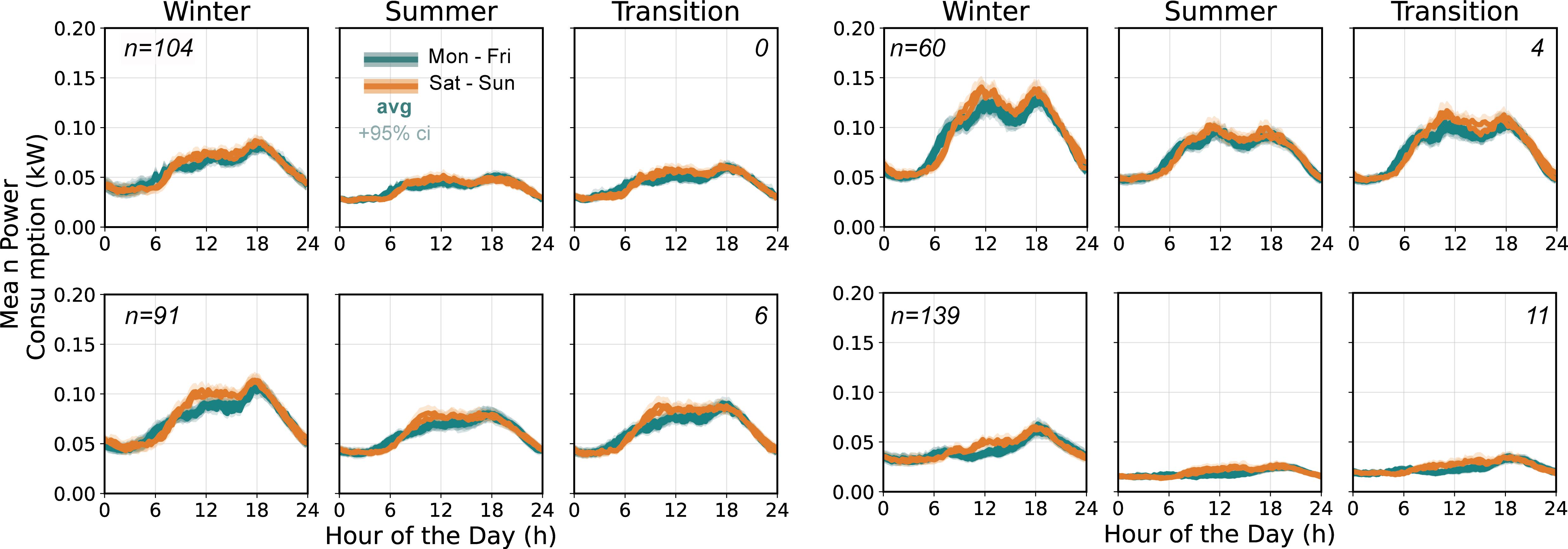}}
\caption{Classification of OM consumers - Shown are the average typical weeks ($\pm$95\% confidence interval) from the top four most populated consumer clusters covering 71\% of OM power consumers (PCA with 5 PCs, k=15, k-means+).}
\label{fig:figure_5}
\end{figure}

\begin{figure}[tbp]
    \centering
    \begin{subfigure}[t]{0.48\textwidth}
        \includegraphics[width=\textwidth, trim=20 20 10 20, clip]{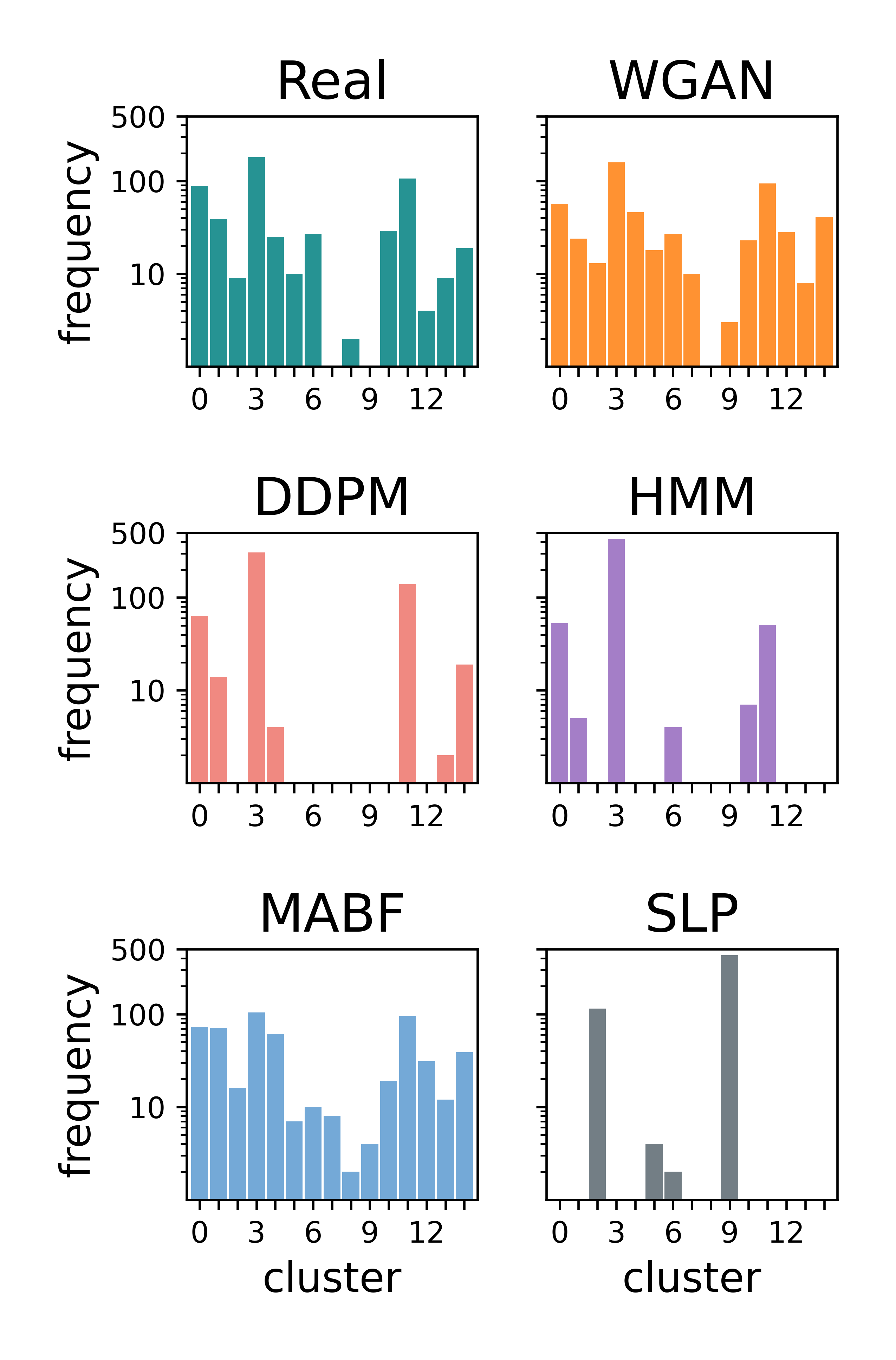}
        \caption{}
        \label{fig:clust_single_original}
    \end{subfigure}
    \hfill
    \begin{subfigure}[t]{0.48\textwidth}
        \includegraphics[width=\textwidth, trim=10 10 10 10, clip]{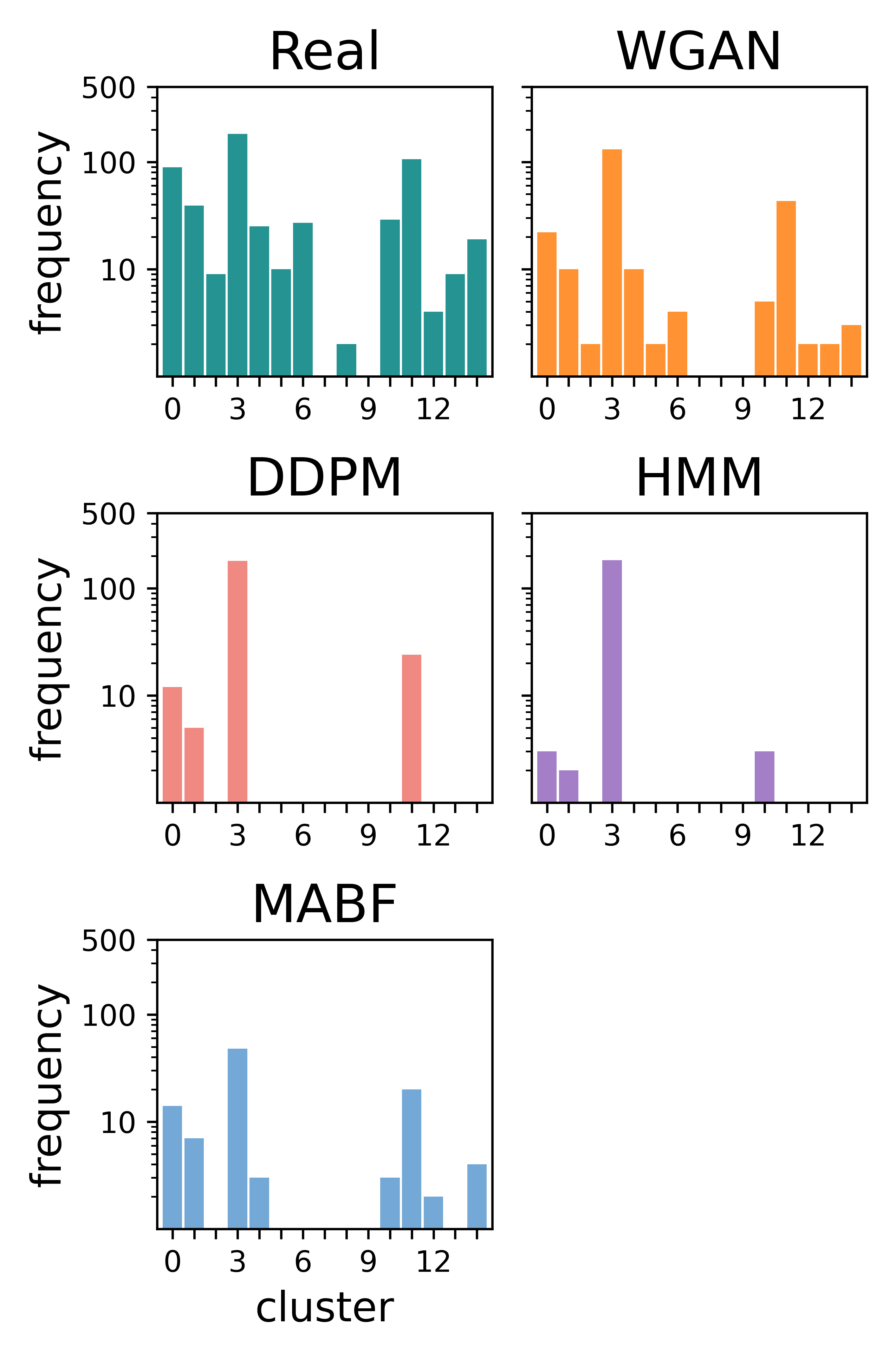}
        \caption{}
        \label{fig:clust_distr_to_paired_original}
    \end{subfigure}
    \caption{Consumer-type distribution - Note: y-axis is log-scaled. a) Initial clustering with \gls{om} data alone (PCA with 5 PCs, k=15, k-means+).  Synthetic profiles were matched to closest centroid. b) Consumer-type preservation of surrogate power consumption data (only original data in clustering, PCA with 5 PCs, k=15). Frequency of clusters in which synthetic power profiles occupy the same cluster as their corresponding original power consumer. Note that SLP profiles did not match any paired original profile.}
\end{figure}

Naturally consumers fall into different power consumption categories due to their individual consumption patterns. Thus, consumer typing was performed on the \gls{om} data to identify different types of consumers (Figure \ref{fig:clust_OM}). Classification of power consumers revealed fifteen stable consumer types, whereby 54\% of consumers are found in 3 major clusters (Figure \ref{fig:clust_single_original}). The clustering sensitivity was high enough to categorize four visually quite particular consumers as an individual type. (Figure \ref{fig:figure_5},  cluster 0,4,6,11).
Next, we used the centroids from the consumer typing of the \gls{om} data and assigned the surrogate power profiles from our various ML-models to the respective consumer types (Figure \ref{fig:clust_all_on_OM}). The digital surrogates from the \gls{WGAN} and \gls{MABF} exhibited a distribution of consumer types that was very close to the \gls{om} data (Figure \ref{fig:clust_single_original}). 

The \gls{HMM}- and \gls{DDPM}-model exhibited a bias as effectively these ML-models  predominantly generated surrogates for the most populated clusters of the \gls{om} data (Figure \ref{fig:clust_single_original}). At last \gls{SLP} surrogates almost exclusively populated two sparse clusters from the \gls{om} data and thus created two pseudo-clusters (Figure \ref{fig:clust_single_original}).

Observing similar consumer-type distributions we analyzed the frequency of a digital surrogate representing its respective \gls{om}-paired consumer of the same consumer type (Figure \ref{fig:clust_distr_to_paired_original}). The \gls{WGAN} created the most correctly assigned surrogates and was able to achieve this for a multitude of clusters with the most populated \gls{om} consumer types being better represented. Similarily, the \gls{MABF} was able to create surrogates correctly for the dominant consumer types (Figure \ref{fig:clust_distr_to_paired_original}). Both \gls{DDPM} and \gls{HMM} had a strong bias in generating power profiles that represented households from the most abundant consumer types (Figure \ref{fig:clust_distr_to_paired_original}).

\begin{figure}[!ht]
\centering
\resizebox{1.0 \textwidth}{!}{\includegraphics{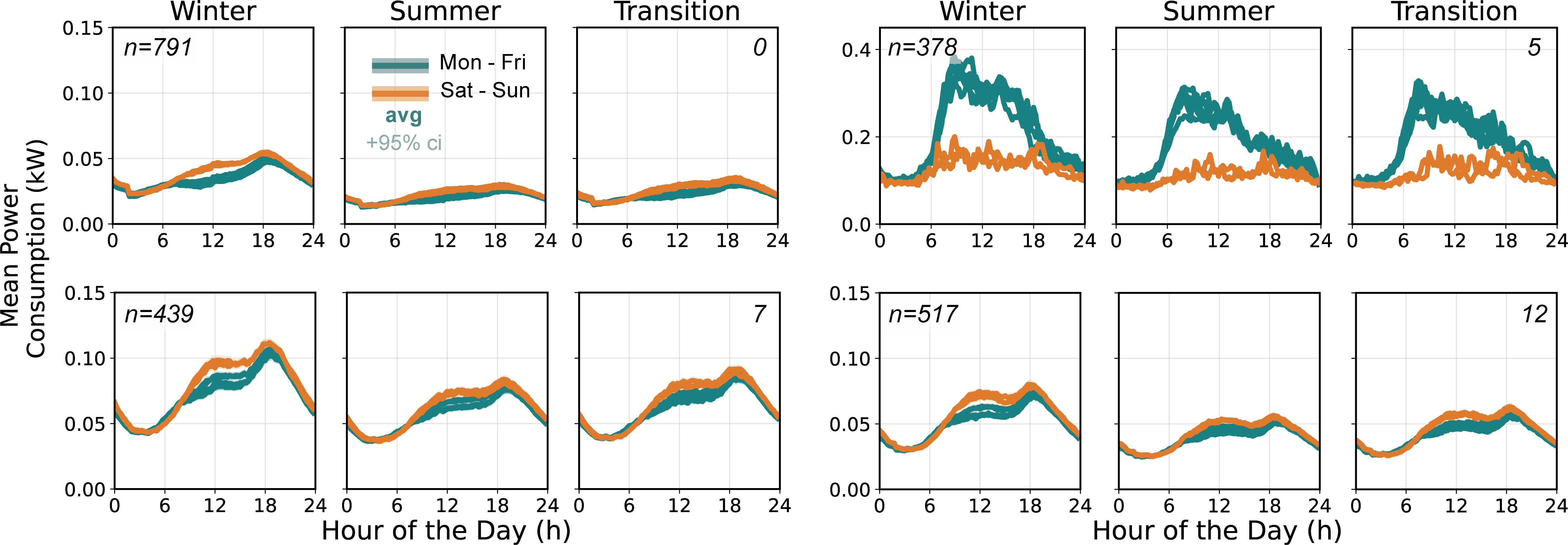}}
\caption{Mixed consumer typing - Shown are the average typical weeks (+ 95\% CI) for the four most populated consumer types (clustering with all power profiles, PCA with 5 PCs, k=15, k-means+).}
\label{fig:clust_all_top4}
\end{figure}

\begin{figure}[tbp]
    \centering
    \begin{subfigure}[t]{0.48\textwidth}
        \includegraphics[width=\textwidth, trim=0 0 0 0, clip]{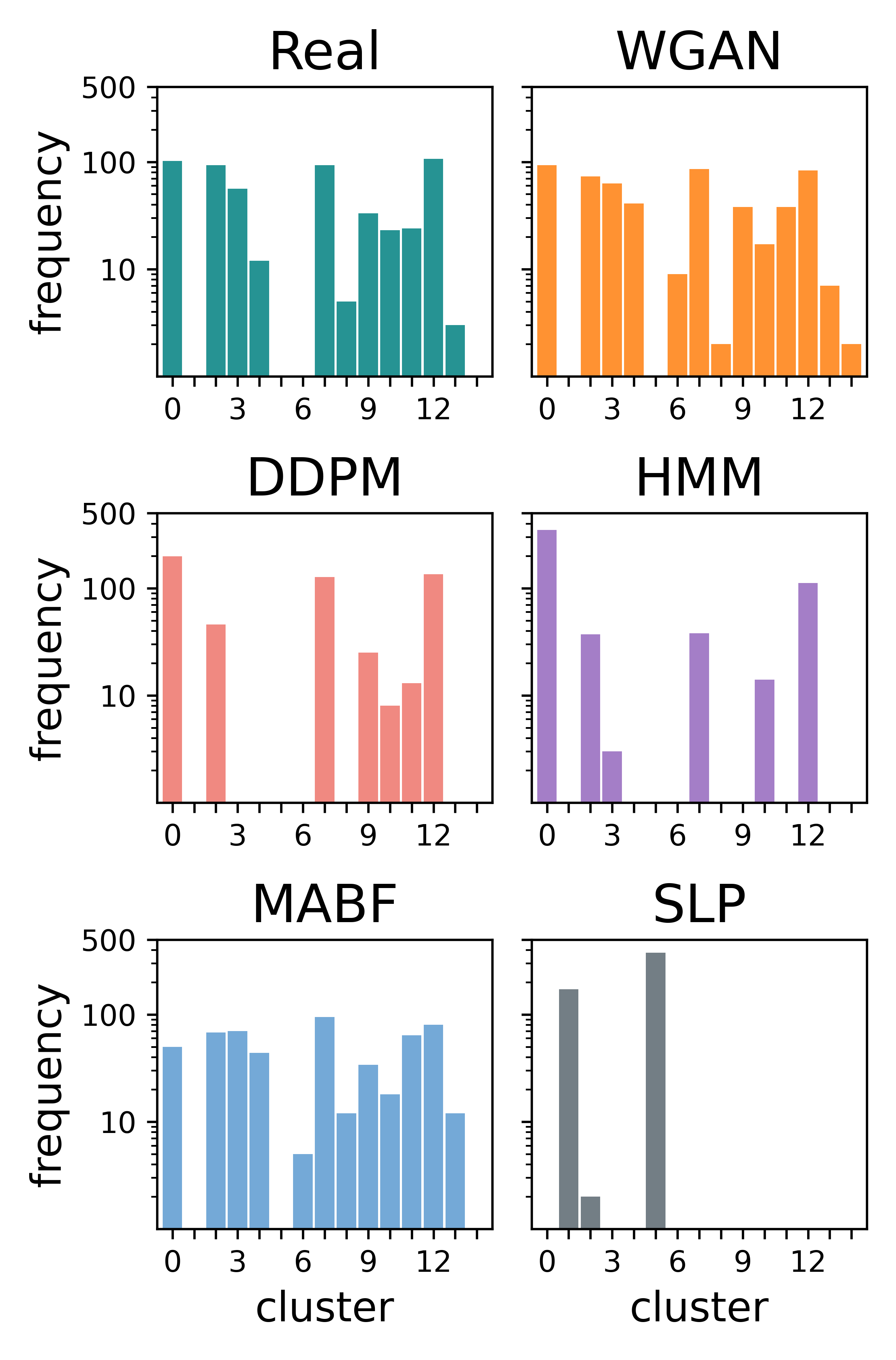}
        \caption{}
        \label{fig:clust_all_data}
    \end{subfigure}
    \hfill
    \begin{subfigure}[t]{0.48\textwidth}
        \includegraphics[width=\textwidth, trim=0 0 0 0, clip]{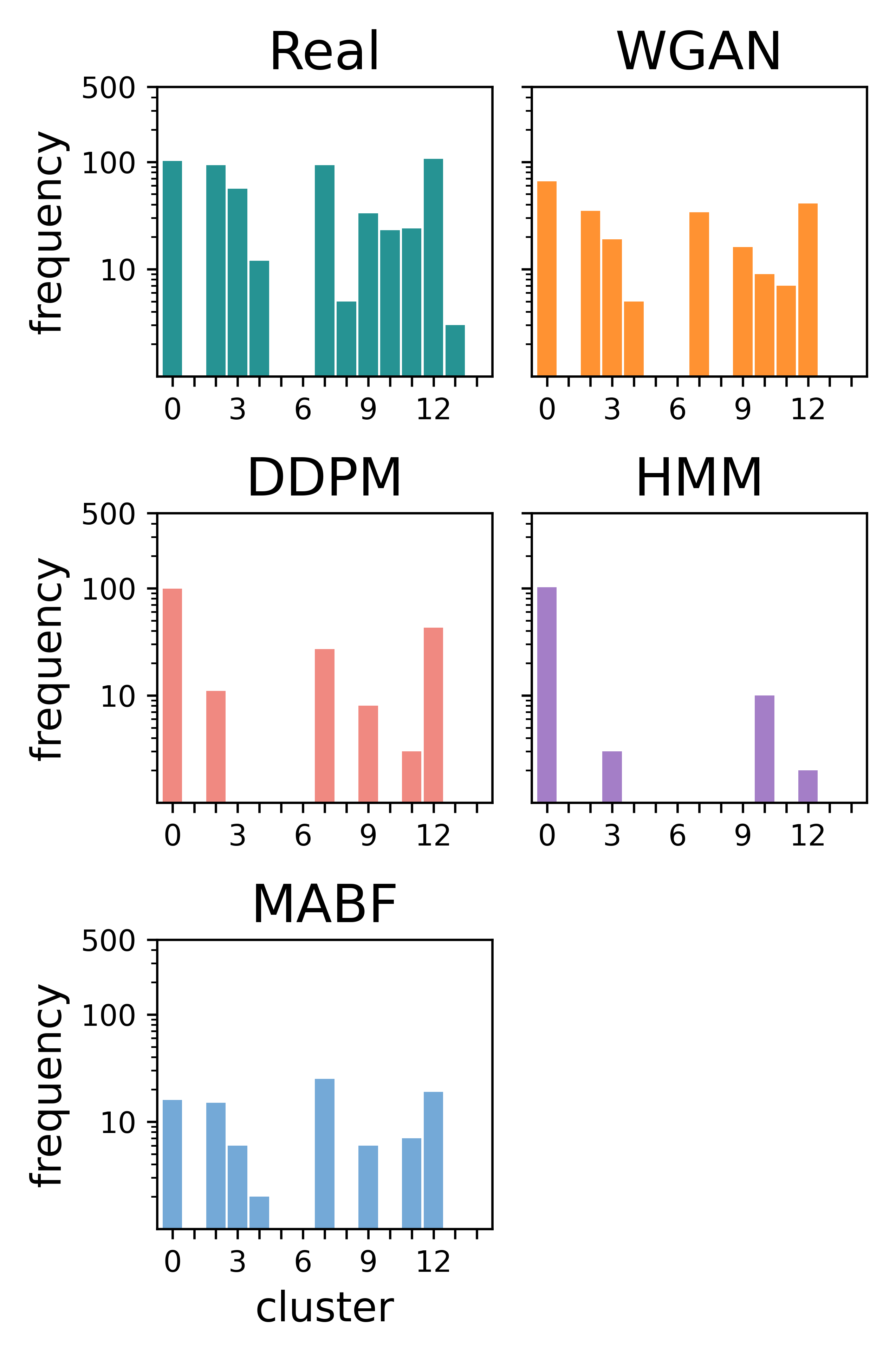}
        \caption{}
        \label{fig:clust_fit_to_original}
    \end{subfigure}
    \caption{Consumer-type distributions - Clustering with all power profiles (all models, PCA with 5 PCs, k=15, k-means+). a) Population of the clusters by the respective model. \textit{Note: SLP-profiles create and populate cluster 1 and 5 exclusively}. b) Frequency of clusters, where synthetic profiles populate the same cluster as their matching original power consumer. \textit{Note: SLP-profiles create their own clusters and never match to original profiles, thus SLP-data is absent here.}}
\end{figure}

Subsequently, we examined the impact of blending \gls{om} power profiles with digital surrogates generated by our ML models on consumer typing. To achieve this, power profiles from all ML models, along with the original data, were clustered together in a unified approach (Figures \ref{fig:clust_all_top4}, \ref{fig:clust_all_blended}). The \gls{SLP} formed and predominantly populated distinct clusters that remained virtually unoccupied by \gls{om} consumers or the other tested ML models (Figure  \ref{fig:clust_all_data}). Overall the consumers were distributed more heterogeneously over four dominant clusters (Figure \ref{fig:clust_all_top4}). The \gls{MABF} and \gls{WGAN} again generated power profiles for all consumer types identified earlier. The \gls{HMM} and \gls{DDPM} however, both had the tendency to learn characteristics from highly abundant consumer types but rarely generated power profiles for sparsely populated \gls{om} clusters (Figure \ref{fig:clust_all_data}).

In addition, we analysed whether digital surrogates populate the same clusters as their paired real power consumer (Figure \ref{fig:clust_fit_to_original}). Again the surrogates from the \gls{WGAN} had the highest frequency in matching their respective original consumer and achieved this for practically all clusters populated by the \gls{om} consumers. However, more populated clusters where matched more often and usually $\le$50\% of digital surrogates represented the same consumer type as their paired original consumer (Figure \ref{fig:clust_fit_to_original}).

Similarily, the \gls{MABF} was able to generate similar consumer types, but matching rate was even more decreased. The \gls{DDPM} and \gls{HMM} on the other hand again exhibited a strong preference to generate surrogates for the highly populated clusters. A consequence of this strong bias for a few consumer types, in the case of the \gls{HMM} effectively one, is the artifact that their surrogate-matching covered all \gls{om} consumers for this type (Figure \ref{fig:clust_fit_to_original}).

\subsection{Correlation Analysis, HMM-State Distributions}
As demonstrated earlier, the tested ML models successfully generated power profiles for specific consumer types, capturing the majority of single-day samples present in the \gls{om} data, as confirmed by UMAP analysis and consumer typing. Furthermore, we explored whether single consumers could be reverse identified. To this end, we extracted various statistical metrics from the time-series data and performed a correlation analysis. Additionally, an HMM model was constructed for each power consumption time series to examine the distributions of identified states, which are typically unique to individual customers.

\begin{figure}[!ht]
\centering
\resizebox{.95\textwidth}{!}{\includegraphics{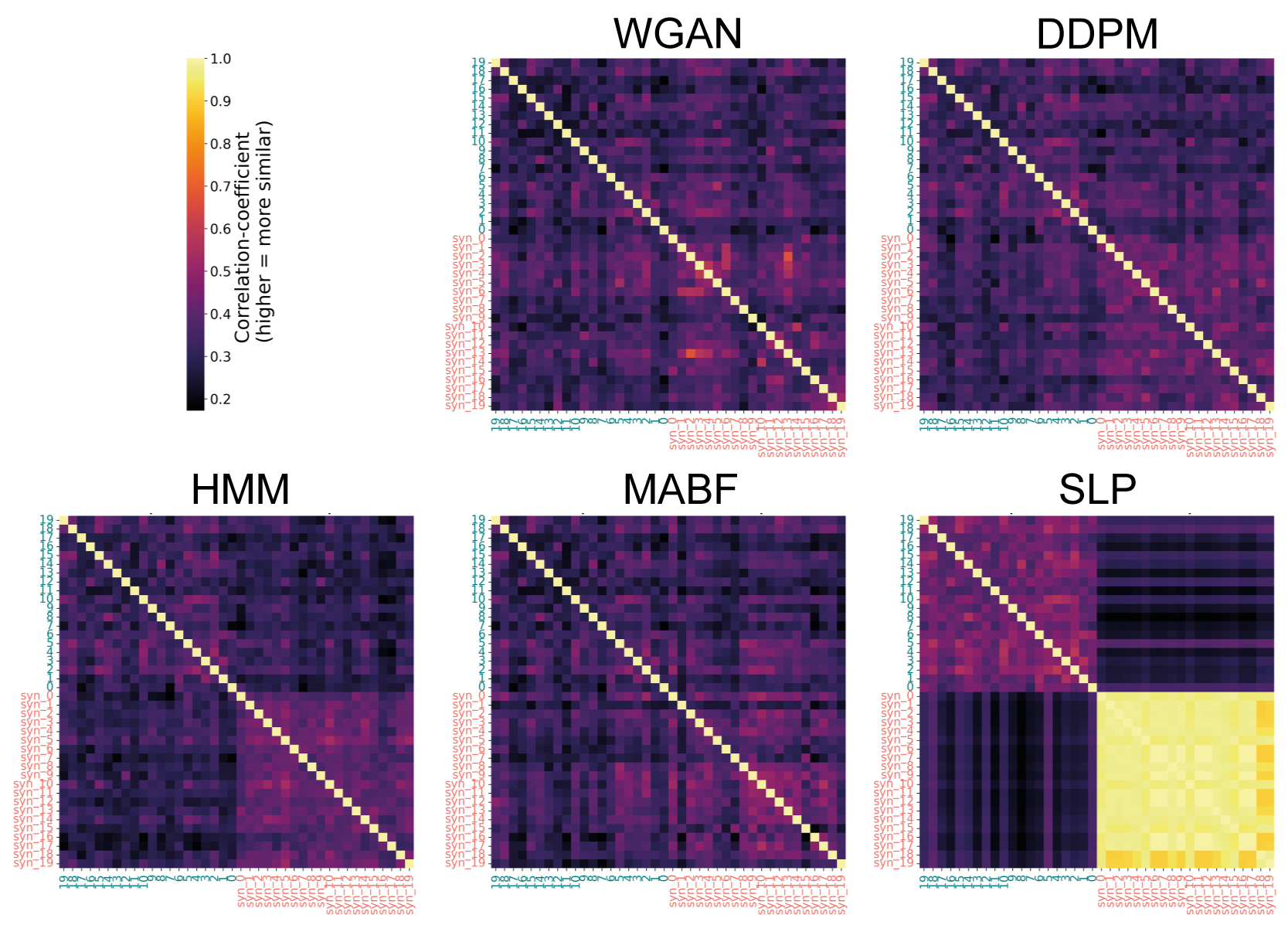}}
\caption{Similarity analysis of power profiles - Correlation analysis of power consumption metrics between real and synthesized power profiles. A higher correlation coefficient indicates higher similiarity. The small matrix segment highlighting only 19 real profiles and their paired synthetic power profile. Note: The SLP power profiles are almost identical to one another and show only very low similarity to real power consumers.
}
\label{fig:corr_analysis}
\end{figure}

Over 2,000 statistical parameters were normalized and included in the correlation analysis. The correlogram revealed no prominent cross-patterns that would be indicative for a high similarity between digital surrogates and their paired original consumer time series (Figure \ref{fig:corr_analysis}). Each ML model exhibited distinct patterns, with certain time series showing higher correlation coefficients, suggesting potential similarities with consumers of the same type, based on shared characteristics in energy consumption (Figure \ref{fig:corr_analysis}). An exception was observed in the \gls{SLP}-generated consumer power profiles, which demonstrated very high internal correlation but minimal similarity to their respective \gls{om} counterparts, consistent with earlier findings (Figure \ref{fig:corr_analysis}) and the stereotypical nature of this model generating power consumption data.

Since the digital surrogates were not direct analogues, a similarity index was calculated for each power consumption time series (Figure \ref{fig:corr_analysis}). This analysis encompassed all consumers generated by a given ML model as well as the \gls{om} consumers. The results revealed that the similarity between a specific \gls{om} consumer and its corresponding surrogate was generally low (Figure \ref{fig:corr_analysis}). As anticipated, the \gls{HMM}-generated data exhibited a sharp elbow in the similarity index, reflecting the bias in generated consumer types. Consistent with earlier findings, the \gls{SLP} profiles, which formed distinct consumer types and displayed very high internal correlation, showed minimal similarity to the \gls{om} data (Figure \ref{fig:corr_analysis}).

Given the overall low correlation scores, we next analyzed the most abundant consumer type among the top 10 most highly correlated \gls{om} consumers for each surrogate. Surprisingly, with the exception of the \gls{SLP}, all models (\gls{WGAN}, \gls{DDPM}, \gls{MABF}, and \gls{HMM}) generated power consumption data that correlated with consumers across all types, with a slight preference for more populated clusters (Figure \ref{fig:corr_analysis}). This indicates that, despite the relatively low separation quality in consumer typing, the extracted statistical parameters are robust for correlation analysis, allowing a quality assessment of the digital surrogates.

While none of the ML models generated perfect analogues to the \gls{om} data, they successfully captured the unique characteristics of the consumer types, resulting in synthetic power consumption data with high statistical similarity to these types.

\subsection{Error and similarity metrics}

Aside from visual diagnostics, consumer typing and a correlation analysis of the digital surrogates a set of standard error metrics (as suggested by \cite{review_scheidt_2020}) was applied to quantify the similarity of the digital surrogates to their respective real-world counterparts (Figure \ref{fig:easy_stats}, Table \ref{tab:error_stats}).

Consistent with earlier findings, \gls{WGAN} and \gls{MABF} models achieved the best performance across all error (Table \ref{tab:error_stats}) and similarity metrics—showing the lowest MAE, MAPE, RMSE, and MMD, and the highest Pearson correlation and SSIM (Figure \ref{fig:easy_stats}, \ref{fig:error-metrics}). In contrast, the \gls{SLP} performed worst in all tested error and similarity metrics whereas \gls{DDPM} and \gls{HMM} exhibited moderate performance (Figure \ref{fig:error-metrics}). Overall, all ML models outperformed the \gls{SLP} in generating power consumption profiles.

\section{Discussion}

Our comparative evaluation of various machine learning models for generating synthetic power consumption profiles reveals significant insights into their respective performance, strengths and limitations in replicating real-world energy consumption patterns. 

First, advanced generative models such as \gls{WGAN} and \gls{DDPM} outperform traditionally used approaches like \glspl{SLP} in terms of replicating temporal dynamics and seasonal variations in energy consumption. However, while the \gls{SLP} serves as a useful baseline, its inability to mimic unique patterns of daily and seasonal energy consumption underlines the need for more sophisticated models.

Among the evaluated models in this work, the \gls{WGAN} exhibited the highest fidelity in capturing daily patterns, seasonal changes and different unique characteristics for very heterogeneous consumer types. Though, the \gls{WGAN} had the tendency to overestimate evening peak-power consumption. Likewise the \gls{DDPM} was able to reproduce seasonal changes in power consumption but generally overestimated power consumption and exhibited a bias for generating surrogates for the most populous consumer types.

The \gls{MABF} demonstrated its potential in handling the complex patterns in power consumption and variabilities between different consumer types, although it also showed a behaviour in overestimating peak-load times of the consumers. In contrast, \gls{HMM}s struggled in reproducing daily fluctuations in power consumption and thus limits its use in generating highly realistic power consumption data with the given input parameters. Nevertheless, the simplicity of \glspl{HMM} and their easy interpretation provide a powerful tool for analytical applications, particularly in the quality assessment of synthetic power consumption data.

\section{Conclusion}
The availability of realistic electric load data is crucial for various applications, such as grid planing and state estimation. Currently most network operators have only limited insight into actual real-time behaviour of the power grid system, highlighting the need for improved data accessibility. However, privacy concerns and GDPR compliance make the use of real data challenging. Additionally, many applications require long-term data to be useful.

The use of digital surrogates, anonymised synthetic power consumption data, provide a valuable alternative to real-world data in these applications. Here, we provide a comprehensive framework to generate digital surrogates of real-world consumers in Germany and evaluate the quality of the synthetic data. The aim was to generate surrogates in a particularly difficult long-term scenario that all models had to master.
For the first time, a German energy prosumer dataset with around 550 households across the country was utilized. Though, only a small number of households and a relatively short time-span of 2.5 years, the dataset offers valuable insights and highlights the need for larger datasets and longer time frames of data. The curated datasets of training and synthetic data generated in this work are also readily available to be used in other applications \cite{sylaski_datarepo_2024}).

Furthermore, while some metadata, such as weather information, is included, integrating more complex socio-economic factors \cite{jiao_2022} is critical for improving modeling accuracy and forecasting performance and to selectively generate power profiles for specific scenarios by entering conditions. Particularly with the rise of "New Power Consumers" (NPCs), such as electric heating or heat pumps, electric vehicles (EVs), or battery storage will provide even more heterogeneity among prosumers and might cause new challenges for ML models in the future.

Assessing the quality of our generated synthetic power consumption data poses a multitude of challenges. We employed a comprehensive approach, utilizing metrics ranging from simple visual diagnostics and dimensionality-reduced data analysis to distribution-based metrics and correlation analysis of extracted statistical characteristics. 

The evaluated models demonstrated their ability to generate digital surrogates that replicate the behavior of real households while maintaining sufficient variability to ensure hitherto GDPR-compliant level of anonymization and are readily usable for applications like state estimations. The trained models can easily be applied to generate power consumption data for any desired time-spans, without temporal constraints.

Future research should implement the latest advances in machine learning like Hybrid-Transformers or federated learning \cite{li_review_federatedlearning_2020, mammen_federated_2021} and should include more continuous and categorial data for even more tailored inferencing for specific scenarios.

In conclusion, this work provides a a significant step in advancing an adaptive framework for long-term energy forecasting by combining high-quality datasets, flexible ML-models and comprehensive evaluation techniques.

\newpage
\section{Appendix}
\section{Methods}

\subsection{Consumer clustering}
Consumer clustering is performed as described recently \cite{riedl2024cigre}. In brief, a PCA is performed on typical weeks, followed by k-means$^{++}$ clustering (k=15) of PC1-5. The average typical week is then calculated for each cluster from the assigned load profiles and plotted.

\subsection{Data Evaluation}
\subsubsection{Error-Metrics}
\begin{itemize}
    \item $ \text{MAE} = \frac{1}{n} \sum_{i=1}^{n} | y_i - \hat{y}_i | $
    \item $\text{MAPE} = \frac{100\%}{n} \sum_{i=1}^{n} \left| \frac{y_i - \hat{y}_i}{y_i} \right|$
    \item $ \text{RMSE} = \sqrt{\frac{1}{n} \sum_{i=1}^{n} (y_i - \hat{y}_i)^2}$
    \item $\text{Pearson correlation} = \frac{\sum_{i=1}^{n} (y_i - \bar{y}) (\hat{y}_i - \bar{\hat{y}})}
{\sqrt{\sum_{i=1}^{n} (y_i - \bar{y})^2} \cdot \sqrt{\sum_{i=1}^{n} (\hat{y}_i - \bar{\hat{y}})^2}}
$
    \item $SSIM(X, Y) = \frac{(2\mu_X \mu_Y + C_1)(2\sigma_{XY} + C_2)}
{(\mu_X^2 + \mu_Y^2 + C_1)(\sigma_X^2 + \sigma_Y^2 + C_2)}$
    \item $\text{MMD}^2 (X, Y) = \mathbb{E}_{x, x' \sim P} [ k(x, x') ] 
+ \mathbb{E}_{y, y' \sim Q} [ k(y, y') ] 
- 2 \mathbb{E}_{x \sim P, y \sim Q} [ k(x, y) ]$

\end{itemize}

\subsubsection{Correlation analysis, -index}
Statistical features are extracted from the load profiles as described by \cite{riedl2024cigre}. Features from the synthetic load profiles are then compared pair-wise with the original data for the testing period. The Pearson correlation coefficient is then plotted in a matrix. Next, the coefficients are sorted descending for each synthetic load profile. Then the index position is determined for the matching paired-original profile and the results are plotted. To score how close the synthetic load profiles are to their original counterpart the mean average precision (MAP) was calculated \cite{manning_introduction_2008}.

Additionally, the most prominent cluster among the top 11 most similar load profiles for a given synthetic profile was determined and the distribution of the most prominent clusters found was then plotted in a histogram and overlayed with the cluster-distribution of the OM data. 

\subsubsection{Uniform Manifold Approximation and Projection (UMAP)}
In order to visualize the overlap of real and synthetic power consumption data a two-dimensional \gls{umap} projection of daily power consumption profiles was generated with data from all tested models and real data \cite{mcinnes1802umap}. \gls{umap} was configured with the parameters: $n_{components}=2$, $n_{neighbours}=10$, $min_{distance}=1$, $metric=euclidean$. 
Settings were chosen from pilot experiments optimizing separation and overlap visualization of real and synthetic samples. Each point represents a single day from the respective model embedded from the high-dimensional feature space.\\
Subsequently similarity was quantified between real and synthetic day samples. For this, a Hungarian matching algorithm was applied to find the closest pairs of day samples between real data and the respective ML model in the UMAP-embedded space. Pairwise Euclidean distances were calculated between all real and synthetic samples of a respective ML model. The Hungarian algorithm was then used to minimize the overall distance to identify and optimal one-to-on matching.
\subsection{Embedding}
\label{sec:Embedding}
\begin{figure}
    \center
    \includegraphics[width=0.5\textwidth]{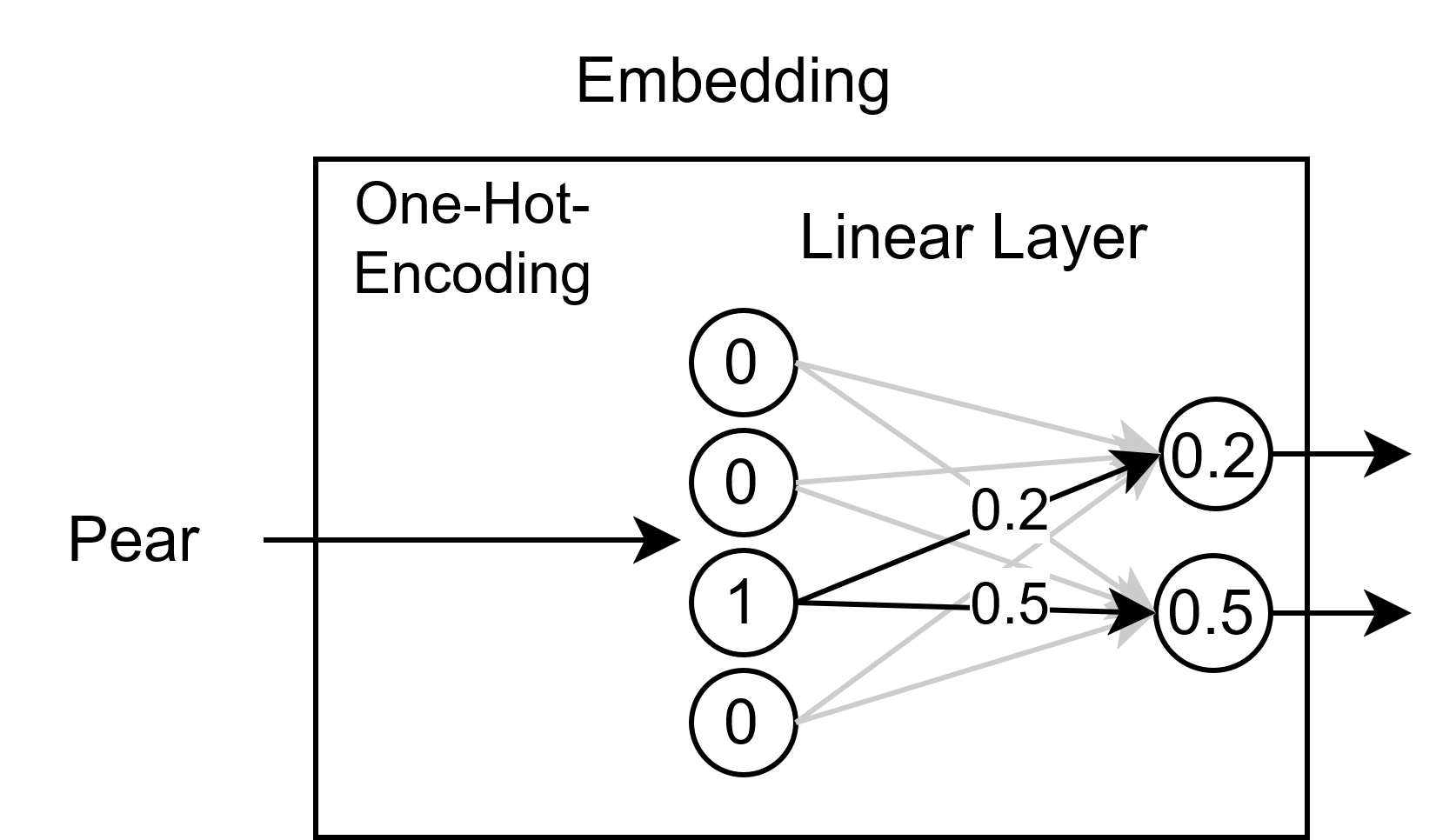}
    \caption{Schematic for an embedding - Example of how an embedding would encode Pear as a two dimensional vector. The gray weights get multiplied by zero and thus do not contribute to the output.}
    \label{fig:EmbeddingExample}
\end{figure}
\begin{figure}
    \center
    \includegraphics[width=.60\textwidth]{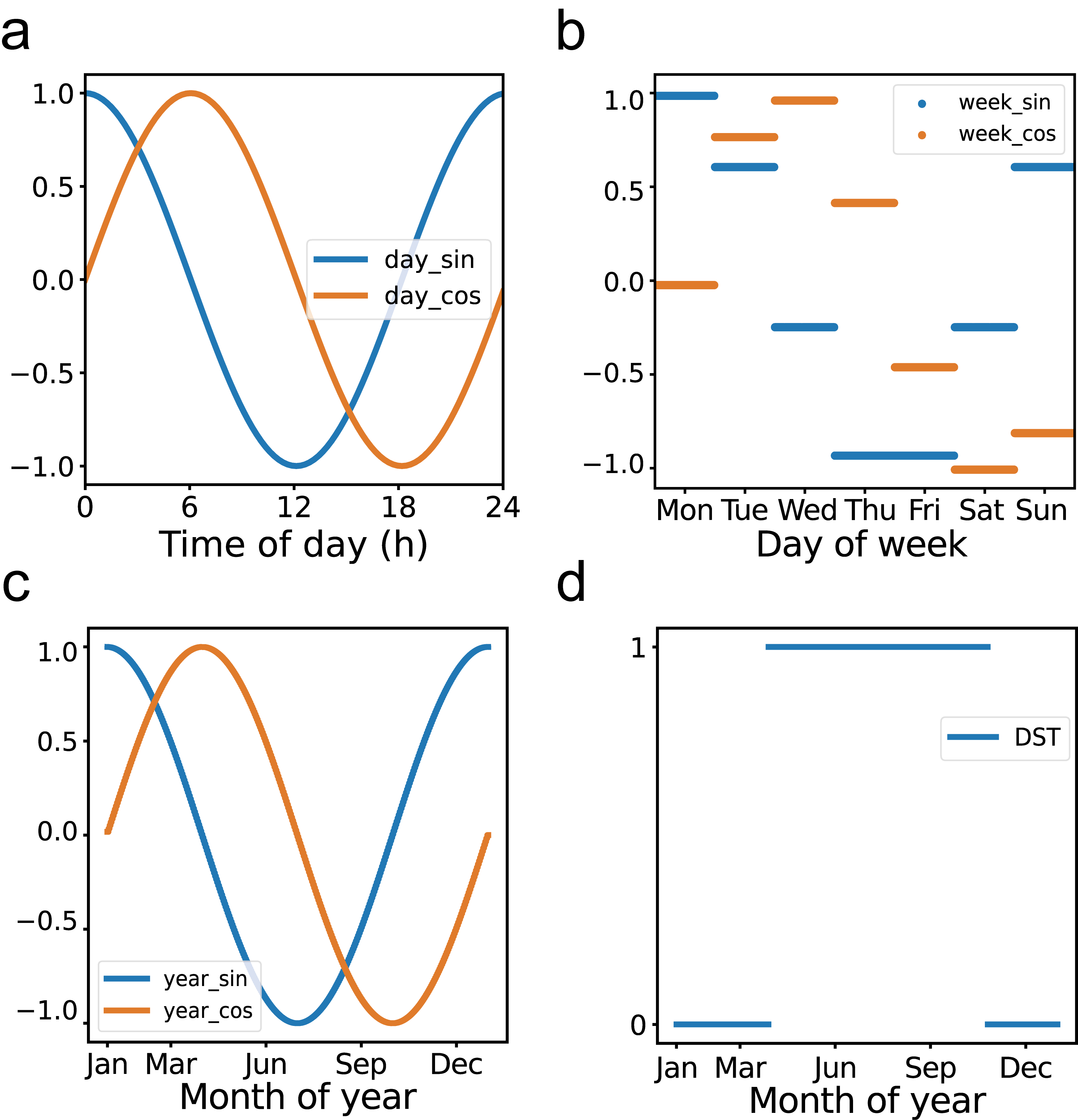}
    \caption{Encoding of time embeddings - a) Data points as sine and cosine over a 24h period b) Encoding of days of the week as sine and cosine c) Encoding of year as sine and cosine d) Daylight-Saving-Time (DST) encoding.}
    \label{fig:time_embeddings}
\end{figure}

An embedding is a method to represent discrete or categorical values as continuous vectors in a higher dimensional space.
The vectors are either hand crafted or the weights of a linear layer on top of the one-hot encoded categorical data \cite{guo2016entity}.
Training this linear layer makes it possible for the model to learn and encode similarities (Figure \ref{fig:EmbeddingExample}) and exploiting these.
Various embedding types were utilized in this manuscript:

\begin{itemize}
    \item \textit{Categorical Embeddings:} Learned vector representations of categorical inputs such as:
    \[
    \ \textit{cluster-id} , \textit{sensor-id}.
    \]Initially, the categorical values are label-encoded and then passed through an embedding layer, which maps each label to a dense, continuous vector of fixed dimension (e.g., \( d \)). For a time window of length \( L \), the resulting tensor of categorical embeddings has shape \( L \times n_{\text{cat}} \), where each entry is a learned vector representation.

    \item  \textit{Time Embeddings (Sine-Cosine):} Deterministic encodings that transform periodic time features (e.g., hour of day, day of week, day of year) into continuous vectors using sine and cosine functions. Time values are first scaled to the range \([0, 2\pi]\) and then transformed as:
    \[
    (\sin_{(\text{time})}, \quad \cos_{(\text{time})}) 
    \]
    where time is either day, week or year, indicating the length of one period, ensuring a smooth and cyclic representation of time. Time embeddings are not learned but are treated as continuous conditions. Additionally daylight saving time is represented as a flag with values one and zero indicating active or inactive daylight saving time. Examples of these embeddings are shown in figure \ref{fig:time_embeddings}.
\end{itemize}

Each embedding is then normalized to be in the interval $[0,1]$.

\subsection{Dataloader}
After splitting the dataset into train and test data and label encoding the categorical data, the time series was further split into training samples, typically spanning one day ($L = 96$, at 15min resolution), starting at midnight. All generated training samples were min-max-scaled together for the power consumption data. The time data is transformed into sine-cosine-embeddings as explained before. The temperature data is standardized and samples containing \textit{NaN}s were dropped. Each training sample is represented as a three-tuple:

\begin{enumerate}
    \item categorical conditions (\textit{cluster-id} and \textit{sensor-id}).
    \item continuous conditions (\textit{time embeddings} and temperature)
    \item power consumption time series
\end{enumerate}

\subsection{Synthesis of power profiles}
This section provides more detailed and in depth information about the models used.
\subsubsection{Hybrid Wasserstein GAN (hybridWGAN)}
The \gls{WGAN} used in this paper is implemented using an LSTM as the generator and a feed forward network as the discriminator.
\begin{figure}
    \center
    \includegraphics[width=\textwidth]{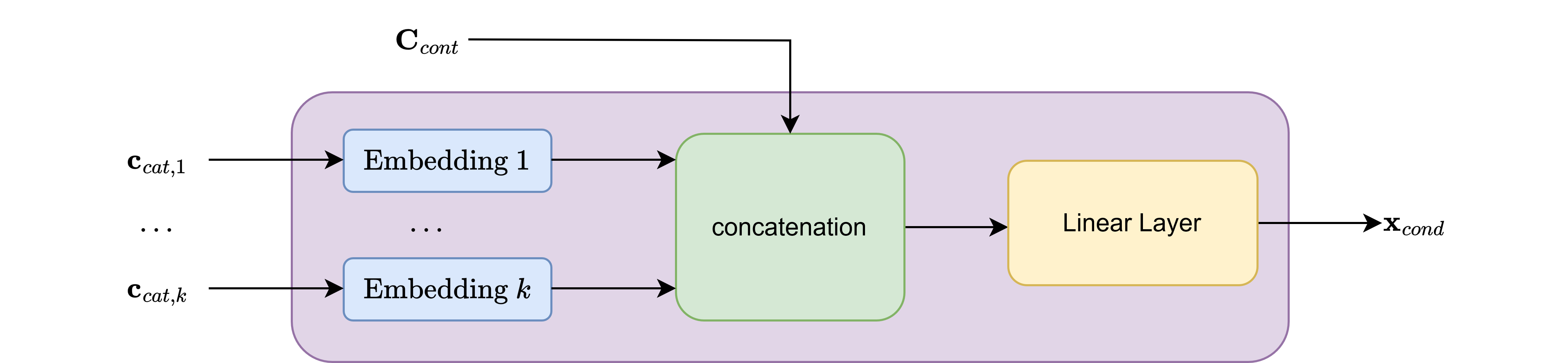}
    \caption{Scheme of the Condition Block}
    \label{fig:conditionalBlockScheme}
\end{figure}
This model is extended by condition blocks (Figure \ref{fig:conditionalBlockScheme}). %

\subsection{Generative adversarial networks (GANs)}
GANs were introduced a decade ago \cite{goodfellow2020generative} and are quite versatile. Usually GANs are widely applied in image generation, image-to-image translation \cite{zhu2017unpaired}, text-to-image generation and data augmentation but can also be used for molecular design and drug discovery [see \cite{lin2020relevant} for a comprehensive review].

At its core GANs are made out of two artificial neural networks (ANNs), a generator and a discriminator network. While the generator is optimized to generate samples that are highly similar to the ground-truth (real input data), the discriminator is trained to weight and classify if a presented sample belongs to the distribution of generated or real data. Both ANNs are trained against each other in an adversarial training procedure to improve the quality of the generated samples.

The generator takes noise, in most cases sampled from a uniform or Gaussian distribution $p_z(z)$, and tries to generate samples $G(z)$.
The objective function of the whole GAN can then be written as: 
\begin{equation}
    \min_{G} \max_{D} V(D,G) = \mathbb{E}_{x\sim p_{data}(x)}[\log D(x)] + \mathbb{E}_{z \sim p_z(z)}[\log(1 - D(G(z)))]
\end{equation}
In practice the righthandside of the sum often leads to vanishing gradients for the generator $G$, as mentioned in \cite{goodfellow2020generative}. 
As one solution for this problem, rather that minimizing $\log(1 - D(G(z)))$ the generator will try to maximize $\log D(G(Z))$. 
While GANs perfom well when trained correctly, there are also drawbacks.\\
One of the most prominent pitfalls is the instable training of GANs if the performance progress of the two networks is not well-balanced. 
If the discriminator is not trained well enough it fails to provide useful feedback to the generator. 
As a result the generator will not be able to learn the distribution $p_{data}(x)$ and is unable to generate realistic samples. 
On the other hand, if the discriminator saturates, the gradients for the generator vanish and the generator is unable to learn. \\
Another problem is modecollapse or the "Helvetica scenario" as it is called in \cite{goodfellow2020generative}.
In that situation, the generator learns to output realistic examples but fails to capture the whole distribution. 
The generator only outputs the same subset of samples, that the discriminator can not identify as fake, with very little variation. 
Thus the generator fulfils the objective function but is not particularly useful as it can't generate the whole spectrum of samples.

\subsubsection{Wasserstein-GAN}
To overcome some of the issues pointed out before Arjovsky et. al. introduced a new loss in \cite{arjovsky2017wasserstein}. 
Instead of optimizing the aforementioned objective function, which effectively minimizes the Jensen-Shannon divergence, they propose to employ the Wasserstein or Earth-Mover distance between the generated and synthetic data.
They argue that the Wasserstein distance provides useful gradients to the generator, even in the case that the discriminator saturates. 
This way the training becomes more stable and less prone to badly tuned hyperparameters or bad initialization.\\
One problem that arises when using the Wasserstein distance instead, is that its infimum is highly untractable. 
By using the Kantorovich-Rubinstein duality \cite{villani2008optimal}, they argue that it becomes tractable, when the discriminator is a 1-Lipschitz function.
To enforce this constraint they use weight clipping. 
In their paper they suggest clamping all weights to the interval $[-c, c]$ (typically with, e.g., $c := 0.01$) after each gradient update. 
They also point out that weight clipping is not a good solution to enforcing the 1-Lipschitz constraint, because it introduces a new hyperparameter that has to be tweaked to the right value for the network to train properly. 
When the parameter is too large, it takes a long time to reach its optimum and if the parameter is too small, the same problem of vanishing gradients can arise again. 
Using the Wasserstein distance leads to the following objective function: 
\begin{equation}
    \min_G \max_{D\in\mathcal{D}} \mathbb{E}_{x\sim p_{data}}[D(x)] - \mathbb{E}_{z\sim p(z)}[D(G(z))]
\end{equation}
where $\mathcal{D}$ is the set of 1-Lipschitz functions. 
The discriminator using this objective function is then often called critic, as it doesn't output a probability anymore but a score from $- \infty$ to $\infty$.
For the sake of simplicity the non-generative part of the GAN will be called discriminator throughout this paper.\\
A better solution to enforce the constraint is proposed in \cite{gulrajani2017improved}. 
In their paper they show that weight clipping for once limits the discriminator to learn only very simple functions.
These functions often are not enough to effectively capture the whole distribution as it leads to a disregard of higher moments of said distribution.
Second, they demonstrate that the right value for the weight clipping is very crucial to keep the gradients from vanishing or exploding. 
As a solution they propose gradient penalty. 
Instead of restraining the weights directly they constrain the gradient norm of the critic's output with respect to its inputs. 
This leads to the following objective function: 
\begin{equation}
    \label{eq:objCrit}
    L = \mathbb{E}_{x\sim p_{data}}[D(x)] - \mathbb{E}_{z\sim p(z)}[D(G(z))] + \lambda \mathbb{E}_{\hat{x} \sim p(\hat{x})}
\end{equation}
$\hat{x}$ is obtained by linearly interpolating pairs of points from the original data $p_{data}$ and the generated data $G(z)$. 
$\lambda$ is the so-called penalty coefficient, a hyperparameter to weight the gradient panelty. 
The authors suggest using $\lambda = 10$ as the standard value, as it showed good results across a wide range of datasets. 
It is worth pointing out that this new objective function is only applied to the discriminator, to provide useful gradients for the generator. 
The generator itself is still trained on the slightly modified objective function originally proposed, that is: 
\begin{equation}
    \label{eq:objGen}
    L = \max_G \log D(G(Z))
\end{equation}
In this paper the algorithm derived in \cite{arjovsky2017wasserstein} using the Wasserstein distance and gradient penalty will be used for GAN training, as this circumvents many of the original problems of GANs. 

\begin{algorithm}[ht!]
    \caption{Training Wasserstein GAN with gradient penalty using the default parameters}
    \label{alg:GANTraining}
    \DontPrintSemicolon
    \SetKwFunction{function}{selectSamples}

    \SetKwInOut{Input}{input}\SetKwInOut{Output}{output}\SetKwInOut{Global}{global}

    \Input{\hspace{0.1cm} $\theta$  \hspace{0.1cm}// \textsl{Inital generator parameters}\\
    \hspace{0.1cm} $w$ \hspace{0.14cm}// \textsl{Inital critic parameters}\\
    }

    \medskip

    \Output{\hspace{0.1cm}  $\theta$ \hspace{0.06cm}// \textsl{Optimal generator parameters}\\
    \hspace{0.1cm}  $w$ \hspace{0.06cm}// \textsl{Optimal critic parameters}
    }

    \medskip

    \Global{\hspace{0.1cm} $\lambda \gets 10$ \hspace{0.1cm}// \textsl{penalty coefficient}\\
    \hspace{0.1cm} $n_{critic} \gets 5$ \hspace{0.1cm}// \textsl{number of critic iterations per generator iteration}\\
    \hspace{0.1cm} $\alpha \gets 0.001$ \hspace{0.1cm}// \textsl{learningrate}\\
    \hspace{0.1cm} $\beta_1 \gets 0.9, \beta_2 \gets 0.999$ \hspace{0.1cm}// \textsl{adam hyperparameters}\\
    \hspace{0.1cm} $m$ \hspace{0.1cm}// \textsl{batchsize}\\
    \hspace{0.1cm} $n_{epochs}$ \hspace{0.1cm}// \textsl{number of epochs to train for}\\
    }
    \medskip

    \;
    \nl \textbf{for} $i = 1, \dots, n_{epochs}$ \textbf{do}\;
    \nl \qquad \textbf{for} $j = 1, \dots, n_{critic}$ \textbf{do}\;
    \nl \qquad \qquad \textbf{for} $k = 1, \dots, m$ \textbf{do}\;
    \nl \qquad \qquad \qquad $x \sim p_{data}$ // sample real data\;
    \nl \qquad \qquad \qquad $z \sim p(z)$ // sample latent variable\;
    \nl \qquad \qquad \qquad $y \gets G(z)$\;
    \nl \qquad \qquad \qquad $\hat{x} \gets \epsilon x + (1 - \epsilon)y$ //interpolate\;
    \nl \qquad \qquad \qquad $L^i \gets D(y) - D(x) + \lambda(\| \nabla_{\hat{x}} D(\hat{x})\|_2 - 1)^2$ // \ref{eq:objCrit}\;
    \nl \qquad \qquad \textbf{end for}\;
    \nl \qquad \qquad $w \gets Adam(\nabla_w \frac{1}{m} \sum^m_{i=1}L^i, w, \alpha, \beta_1, \beta_2)$\;
    \nl \qquad \textbf{end for}\;
    \nl \qquad $\{z^i\}^m_{i=1} \sim p(z)$\;
    \nl \qquad $\theta \gets Adam(\nabla_\theta \frac{1}{m} \sum^m_{i=1} -D_w(G_\theta(z)),\theta, \alpha, \beta_1, \beta_2)$ //\ref{eq:objGen}\;
    \nl \textbf{end for}\;

\end{algorithm}

\begin{figure}
    \center
    \includegraphics[width=\textwidth]{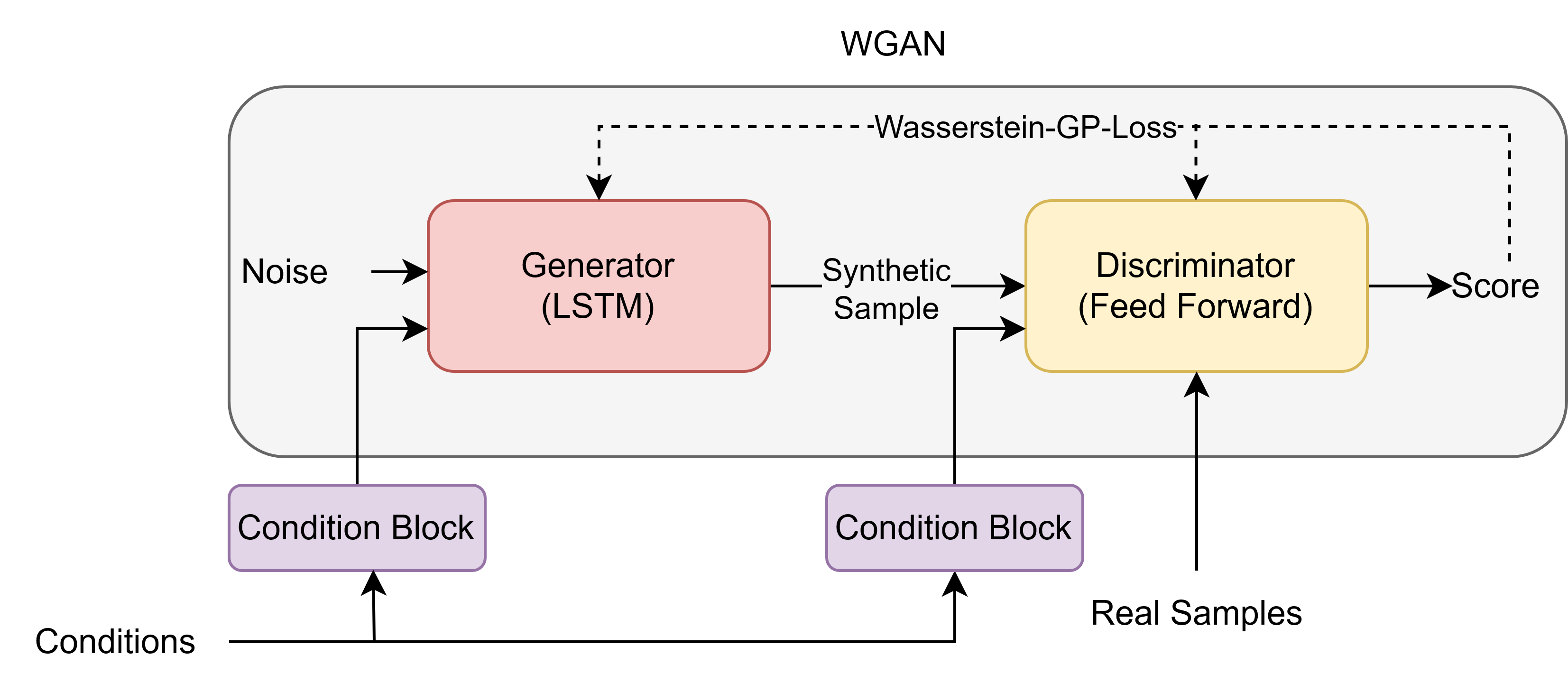}
    \caption{Scheme of the WGAN with conditions}
    \label{fig:CWGAN}
\end{figure}

With this architecture it is possible to generate generic power profiles.
While this architecture can learn to generate a variety of power profiles it lacks the ability to generate specific power profiles with different characteristics.
One example is the dependency between power consumption and day of the week or seasons.
Since the model does not get any information in that regard it cannot directly learn to produce powerprofiles that show characteristics of a Saturday in the winter. It is possible that the model could learn these dependencies when trained on power profiles that span a whole year.
This approach is not sensible for example because it would drastically reduce the number of samples that the model can be trained on.
\subsubsection{Conditional WGAN}
Another approach is to use the aforementioned meta-data that is available in the dataset.
We selected the meta-data categories that are probably most useful for the model to improve realism of the generated power profiles.
The selected categories are the sensor-id, the time of the recording as well as the temperature at that time.
The temperature is determined via the historic data from the ERA5 database given the GPS coordinates of the sensor.
Additionally the cluster-id that the sensor falls into is also used.
With this additional data the model should be able to learn to generate more realistic power profiles for a specific time period, that also shows a behavior similar to that of the sensor.
One challenge with this additional data is representing it in a way that is useful for the model.
For the temperature that is no concern, as it is already represented by numeric values on an interpretable scale.
While the time of the recoding is also in a numeric format, the format does not reflect the structure of the data well.
Since the time is measured in a continuously increasing way two Saturdays are numerically far away, even tho they show very similar power profiles.
The same goes for seasons and other periodicities too.
To circumvent this issue the timestamps are encoded via sine and cosine functions. 
The usage of these functions is motivated by the fact that it is easy to model periodicity and similarity with them. 
Since we want to model daily, weekly and yearly periodicity the time is modeled as 3 sine and 3 cosine functions each, one of each for each periodicity. 
The resulting functions can be seen in figure \ref{fig:time_embeddings}.
The final problem with encoding the time is the daylight saving time. 
The switch between summer to wintertime introduces a shift of one hour between one half of the year and the other half.
As a solution to this a condition for the summer time is introduced in the form of a flag that is 1 for the time where daylight saving time is active and 0 otherwise. 
This flag should solve all these aforementioned problems. 
First it should make the generated time series even more realistic as the model can learn when to shift the generated time series. 
Second it alleviates the preprocessing as the model can just be trained on data with utc time and no preprocessing to account for the daylight saving time should be necessary. 
Another big upside of these conditions is that a model could be trained with time series with different starting times compared to the baseline model where each time series had to be exactly one day and always start at the same time. 
This can make preprocessing even simpler and could improve the training further but examining this is not part of this work.
Lastly the sensor-id and cluster-id have to be represented in a meaningful way.
Since these ids are strings and categorical in nature an embedding is used, to transform the strings into vectors.
The full set of conditions are then processed by the condition block seen in Figure \ref{fig:conditionalBlockScheme}.
Each categorical condition gets its own embedding.
In this case there are two embeddings, one for the sensor-id and one for the cluster-id.
The preprocessed categorical data is transformed into vectors using these embeddings.
The preprocessed continous conditions (time and temperature) are then concatenated together with the embeddings.
The resulting vector is then fed to a linear layer that outputs the final representation of the conditions as a vector that is then fed as an additional input into the WGAN.
A scheme of the resulting model is given in figure \ref{fig:CWGAN}.
The generator and discriminator each get their own condition block, as they receive different amounts of information from each condition. 
For the generator this is straight-forward, since LSTMs can already handle multidimensional inputs by design. 
Since the discriminator also needs the conditions to provide valuable feedback to the generator it also has to get the conditions as input.
While it is in theory possible to also use an LSTM for the discriminator, we used a fully connected Feed-Forward-Network instead.
This architecture cannot handle multidimensional inputs, so the shapes have to be adjusted.
One solution is to flatten the whole resulting condition matrix and consequently to enlargen the input layer of the discriminator, so that it gets exactly the same information as the generator.
A drawback of this approach is that the input to the discriminator is then mostly composed of the conditions and only a small part is comprised of the time series itself.
For example if embed-dim is 64 and the time series have length 96 the flattened condition vector has size $ 64*96=6144$ because there is one condition vector for every time step, which is enormous compared to the length of the time series.
This makes training the discriminator more difficult and will in turn also affect the training of the generator negatively, as the discriminator has to give feedback to the generator.
Another solution is to only pass the conditions for the first time step to the discriminator.
This way the discriminator gets only a fraction of the information that the generator gets, but that could help the discriminator already enough to still provide valuable feedback to the generator.
The upside is that the input space doesn't get bloated nearly as much as with the flattened conditions, making it easier for the discriminator to learn and recognize relevant features for its task.

\subsubsection{DDPM}

\begin{figure}[!ht]
\centering
\resizebox{0.5\textwidth}{!}{\includegraphics{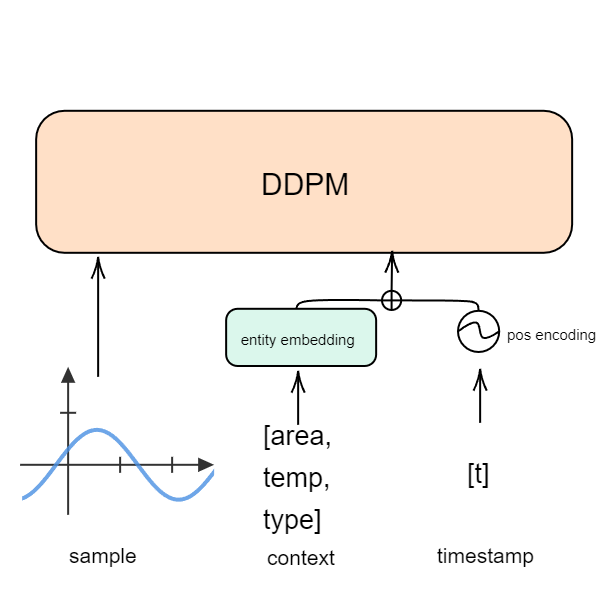}}

\caption{Schematic of the DDPM framework. Inputs are samples, a variety of conditions consisting of contextual conditions and the noising/denoising timestamp $t$ that is positionally encoded.}\label{fig:unet architecture}
\end{figure}

\begin{figure}[!ht]
\centering
\resizebox{0.9\textwidth}{!}{\includegraphics{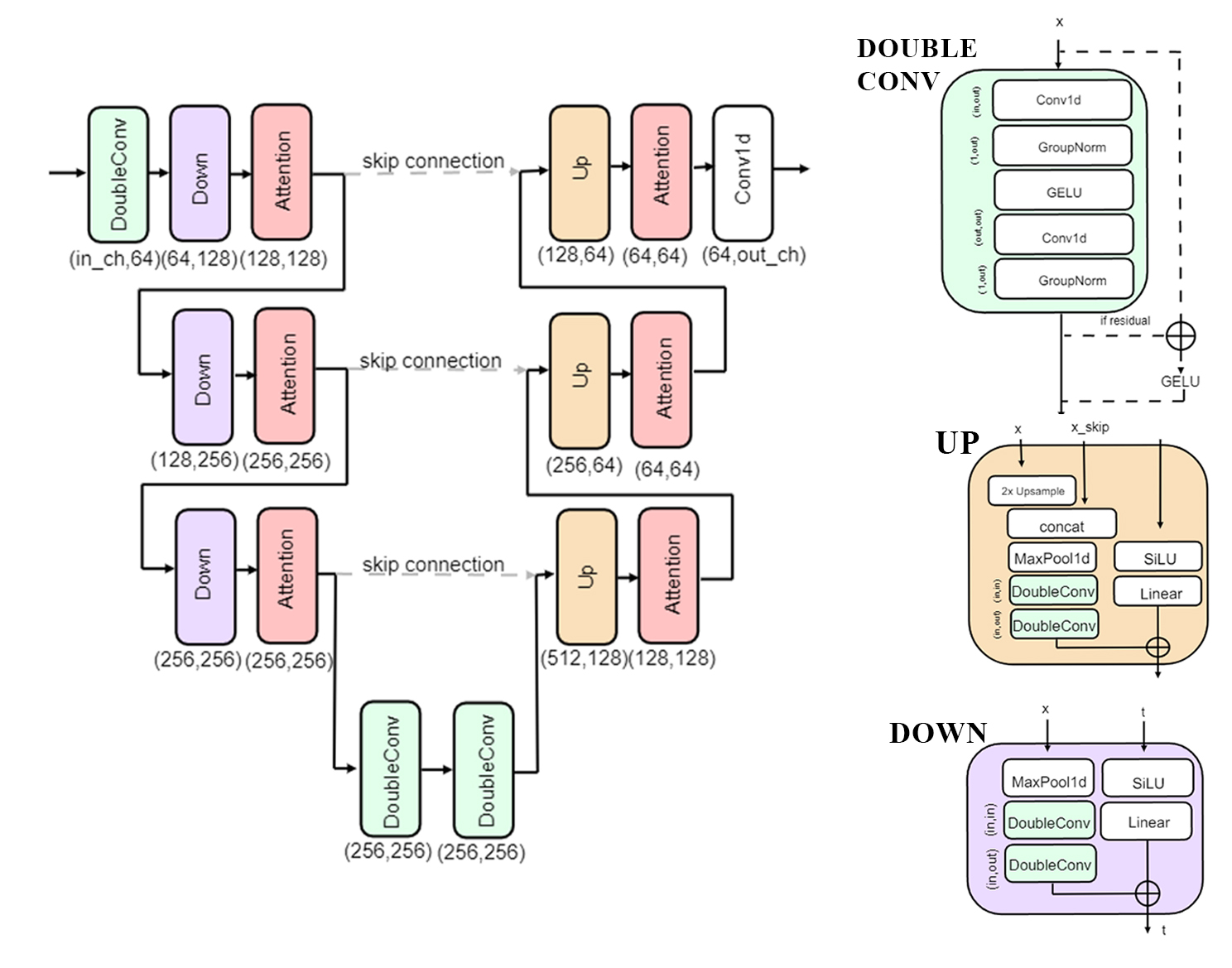}}
\caption{Schematic of the DDPM UNET-architecture and building blocks.}
\label{fig:unet architecture2}
\end{figure}
This process has modified for time series data (Figure \ref{fig:unet architecture}). Creating a dataset that consists of samples of a certain sequence length, leveraging the degrading process to train a noise predictor which can be used in the sampling process is the goal. The underlying architecture used for the noise predictor part in the DDPM is a 1D-UNET mostly resembling the known UNET \cite{ronneberger2015u} architecture, also utilizing self attention layers in between the up- and down sampling blocks (Figure \ref{fig:unet architecture2}).

\subsubsection{MABF}
The proposed \gls{MABF} model leverages recent advancements in \glspl{nf} and \glspl{ctm} research, two distinct yet related approaches to modeling complex probability distributions.
Whether the task is generating probabilistic forecasts or sampling potential future scenarios, it can be formulated as conditional density estimation, aiming to recover the conditional joint distribution $F_{\mathbf{Y}|\mathbf{X}}(\mathbf{y}|\mathbf{x})$ that generated the observed samples $\mathbf{y}_i$ given a set of exogenous variables $\mathbf{x}_i$ for $i \in \{1,\ldots,N\}$ data points.

\Glspl{nf} have gained significant attention in the deep learning community for their ability to model high-dimensional data, such as images and audio~\cite{Kingma2018,Oord2016}, and their application in variational inference~\cite{Rezende2016}.
These models define a sequence of invertible and differentiable transformations to describe a generative process from the base distribution $F_Z$ to the target distribution~\cite{Papamakarios2021,Kobyzev2021}.
To model multivariate distributions, autoregressive models are the most prevalent due to their computational efficiency~\cite{Papamakarios2021}.
\glspl{ctm}, on the other hand, have been widely studied in statistics for modeling various data types, including continuous, discrete, and censored data~\cite{Hothorn2014,Hothorn2018}.
These models define the transformation from the unknown response distribution $F_Y$ to a known base distribution $F_Z$, often using flexible Bernstein polynomials~\cite{Farouki2012} and provide interpretable insights into covariate effects through structured additive predictors~\cite{Kneib2021}.
\glspl{ctm} have recently been combined with neural networks, achieving state-of-the-art performance in density estimation tasks~\cite{Baumann2021,Sick2021,Arpogaus2023,Kook2021}.
\cite{Rugamer2023} proposed semi-parametric autoregressive transformation models based on Bernstein polynomials for probabilistic autoregressive one-step-ahead time-series forecasting in an interpretable manner.

Both \glspl{nf} and \glspl{ctm} utilize the change of variable formula to calculate the density $f_Y$ from the base density $f_Z$ by
\begin{equation}\label{eq:cov}
  f_Y(y) = f_Z\left(h(y)\right) \left|\det\nabla{h}(y)\right|,
\end{equation}
where the absolute Jacobian determinant $\left|\det\nabla{h}(y)\right|$ normalizes the change in density induced by the transformation, to ensure it integrates to $1$.
To draw samples from $F_Y$, the inverse transformation $h^{-1}$ is required, hence $h$ must be both differentiable and invertible.
Equation \ref{eq:cov} is used to calculate the \gls{nll}, to perform maximum likelihood estimation with state-of-the-art optimization algorithms like Adam~\cite{Kingma2017a}.
Building upon these concepts, we combine the flexibility of autoregressive flows for capturing complex dependencies in multivariate time series with the flexibility and stability of Bernstein polynomials for transformation function approximation.

\emph{Bernstein polynomials} offer several advantages for density estimation due to their excellent numerical stability, ability to uniformly approximate any continuous function on a closed interval, and the ease with which monotonicity constraints can be imposed~\cite{Farouki2012}.
A Bernstein polynomial of order $M$ is defined as:

\begin{equation*}
h(y) = (M+1)^{-1}\sum_{i=0}^M \operatorname{Be}_i(y) \vartheta_i,
\end{equation*}

where $\operatorname{Be}_i(y) = f_{i+1, M-i+1}(y)$ is the density of a Beta distribution with parameters $i+1$ and $M-i+1$ evaluated at $y$, and $\vartheta_0,\ldots,\vartheta_M$ are the Bernstein coefficients.
The monotonicity of $h(y)$ can be guaranteed by enforcing $\vartheta_{i} \leq \vartheta_{i+1}$ for all $i=0,\dots, M-1$, making Bernstein polynomials particularly suitable as transformation functions~\cite{Hothorn2014,Hothorn2018,Sick2021}.

\emph{MAF} leverage the chain rule of probability to factorize a $D$-dimensional conditional joint distribution $f_{\mathbf{Y}|\mathbf{X}}(\mathbf{y}|\mathbf{x})$ into a product of conditional distributions, each of which can be modeled using a flexible transformation function~\cite{Papamakarios2018}.
$$f_{\mathbf{Y}|\mathbf{X}}(\mathbf{y}|\mathbf{x}) = \prod_{i=1}^D f_{Y|\mathbf{X}}(y_i|\mathbf{y}_{<i}, \mathbf{x}),$$
where $\mathbf{y}_{<j} = (y_{1}, \dots, y_{i-1})^\top$ represents all previous values up to dimension $i-1$.
Each conditional distribution $f(y_{i} | \mathbf{y}_{<i}, \mathbf{x})$ is modeled via Equation \ref{eq:cov} using a transformation function $h(y_i|\bm{\vartheta}_i)$.
The parameters $\bm{\vartheta}_i$ of this transformation are controlled by a deep neural network $n( \mathbf{y}_{<i}, \mathbf{x})$, with binary masking inspired by the \gls{MADE} architecture, proposed in \cite{Germain2015}, to enforce the autoregressive property.
While this concept theoretically generalizes to any kind of high-dimensional data, it applies particularly well to time series modeling, where each subsequent value naturally depends on past values~\cite{Sick2025}.

\emph{MABF} for time series modeling, are introduced in the following.
To tackle the problem of \emph{,,synthetic load profile generation''}, we define a \glspl{MAF} with $D=48$ for all 30-minute intervals of a single day and model the transformations with flexible Bernstein polynomials.
We chain three transformations $h=\log \circ h_2\circ h_1$:
(1) a scale term $h_1=\beta\mathbf{y}=\mathbf{z_1}$, to scale data to the domain of the following
(2) flexible Bernstein polynomial $h_2=Be(\mathbf{z_1},\bm{\vartheta})=\mathbf{z_2}$, and
(3) logarithmic transform, to ensure the resulting distribution is fixed to the positive domain.

\begin{figure}
    \centering
    \includegraphics[width=\textwidth]{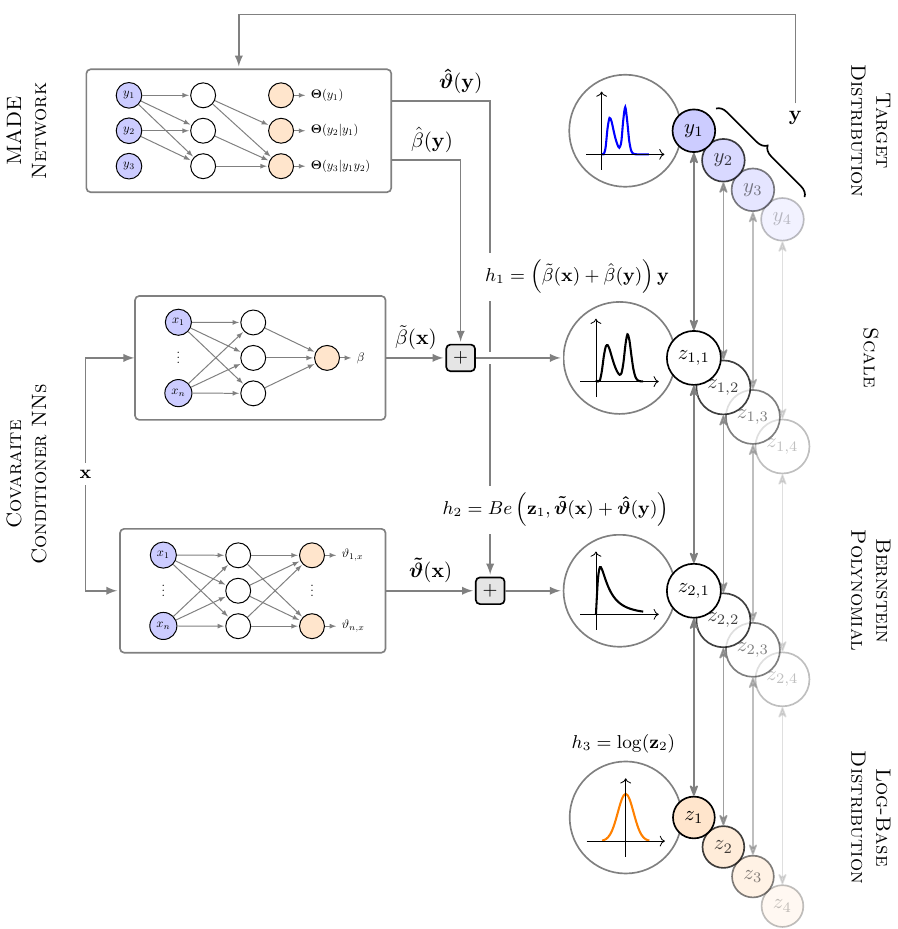}
    \caption{Schematic of the proposed \gls{MABF} model for $D=4$.}
    \label{fig:mabf}
\end{figure}

We use two kinds of neural networks to control the parameters of these chained transformations:
(1) an autoregressive masked neural network to condition on previous values $\hat{\beta},\bm{\hat{\vartheta}}=m(\mathbf{y})$.
(2) two fully connected neural networks $\tilde\beta=n_1(\mathbf{x})$ and $\bm{\tilde\vartheta}=n_2(\mathbf{x})$ to condition on additional covariates like sin/cos encoded calendar variable (day of year, weekday, time), the building ID or the outside temperature.
The scale value $\beta =\tilde\beta + \hat\beta$ and Bernstein coefficients $\bm\vartheta = \bm{\tilde\vartheta} + \bm{\hat\vartheta}$ are then obtained by adding the outputs of the individual neural networks.
The model structure is visualized in Figure \ref{fig:mabf} for $D=4$.

By using Bernstein polynomials as transformation functions within the autoregressive flow framework, we can model complex, non-Gaussian distributions while ensuring the monotonicity of the transformations.
The \gls{MADE} architecture employed for the deep neural network that parameterizes the transformation functions ensures the autoregressive property of the flow to only depend on previous time steps and.

\subsubsection{Hidden Markov models (HHMs)}
Individual HMMs were trained for individual consumer types. Temperature (as $^{\circ}C$; scaled: $temp_{scaled} = (temp + 20) / 40$) and power (as $kWh$) where used as input. A rolling mean (window size= 2h) was applied on the power consumption data. Time features are represented by $\sin_{day}$ and $\cos_{day}$ values and scaled by $(feature_{value})_{scaled} = (feature_{value}+1)/2$. Typical weeks - excluding summer and winter holidays - \cite{riedl2024cigre} are extracted from individual load profiles.

The typical weeks were divided into individual days consisting of 96 values (four measurements per hour). Each 24 hour-sample starts at 5:00 am. The 24 hour sequences were sorted by season and weekday and organized as tensors. Note: Depending on the number of profiles in a cluster and the time spanned by the individual profiles, different numbers of tensors were used in the dictionaries. The python package pomegranate \cite{schreiber2018pomegranate} was used to initialize models with 50 states  each. All tensors of each weekday - season combination were fitted to a model. Thus, 3 x 7 models were fitted in total. Next, new synthetic samples were generated from the models. Samples from the respective weekdays of a season were combined to weeks and weeks to complete years. Last the synthetic profiles were re-normalized.

\subsubsection{Hidden Markov models (HHMs)}
Individual HMMs were trained for individual consumer types. Temperature (as $^{\circ}C$; scaled: $temp_{scaled} = (temp + 20) / 40$) and power (as $kWh$) where used as input. A rolling mean (window size= 2h) was applied on the power consumption data. Time features are represented by $\sin_{day}$ and $\cos_{day}$ values and scaled by $(feature_{value})_{scaled} = (feature_{value}+1)/2$. Typical weeks - excluding summer and winter holidays - \cite{riedl2024cigre} are extracted from individual load profiles.

The typical weeks were divided into individual days consisting of 96 values (four measurements per hour). Each 24 hour-sample starts at 5:00 am. The 24 hour sequences were sorted by season and weekday and organized as tensors. Note: Depending on the number of profiles in a cluster and the time spanned by the individual profiles, different numbers of tensors were used in the dictionaries. The python package pomegranate \cite{schreiber2018pomegranate} was used to initialize models with 50 states  each. All tensors of each weekday - season combination were fitted to a model. Thus, 3 x 7 models were fitted in total. Next, new synthetic samples were generated from the models. Samples from the respective weekdays of a season were combined to weeks and weeks to complete years. Last the synthetic profiles were re-normalized.

\newpage

\newpage










\newpage
Declaration of generative AI and AI-assisted technologies in the writing process:

During the preparation of this work the authors used ChatGPT(GPT-3.5) and ClaudeAI (Claude Sonnet 4) in order to improve the fluency and readability of the text. After using the tools/services, the authors reviewed and edited the content as needed and take full responsibility for the content of the publication.

\bibliographystyle{elsarticle-num} 
\bibliography{bibliography.bib}

\begin{thebibliography}{10}
\expandafter\ifx\csname url\endcsname\relax
  \def\url#1{\texttt{#1}}\fi
\expandafter\ifx\csname urlprefix\endcsname\relax\def\urlprefix{URL }\fi
\expandafter\ifx\csname href\endcsname\relax
  \def\href#1#2{#2} \def\path#1{#1}\fi

\bibitem{review_hong_2020}
T.~Hong, P.~Pinson, Y.~Wang, R.~Weron, D.~Yang, H.~Zareipour,
  \href{https://ieeexplore.ieee.org/document/9218967/}{Energy forecasting: A
  review and outlook}, {IEEE} Open J. Power Energy 7 (2020) 376--388.
\newblock \href {https://doi.org/10.1109/OAJPE.2020.3029979}
  {\path{doi:10.1109/OAJPE.2020.3029979}}.
\newline\urlprefix\url{https://ieeexplore.ieee.org/document/9218967/}

\bibitem{review_scheidt_2020}
F.~v. Scheidt, H.~Medinová, N.~Ludwig, B.~Richter, P.~Staudt, C.~Weinhardt,
  \href{https://www.sciencedirect.com/science/article/pii/S2666546820300094}{Data
  analytics in the electricity sector – a quantitative and qualitative
  literature review}, Energy and {AI} 1 (2020-08-01) 100009.
\newblock \href {https://doi.org/10.1016/j.egyai.2020.100009}
  {\path{doi:10.1016/j.egyai.2020.100009}}.
\newline\urlprefix\url{https://www.sciencedirect.com/science/article/pii/S2666546820300094}

\bibitem{kim_cho_2019}
T.-Y. Kim, S.-B. Cho,
  \href{https://linkinghub.elsevier.com/retrieve/pii/S0360544219311223}{Predicting
  residential energy consumption using {CNN}-{LSTM} neural networks}, Energy
  182 (2019-09) 72--81.
\newblock \href {https://doi.org/10.1016/j.energy.2019.05.230}
  {\path{doi:10.1016/j.energy.2019.05.230}}.
\newline\urlprefix\url{https://linkinghub.elsevier.com/retrieve/pii/S0360544219311223}

\bibitem{wang_lstm_2020}
J.~Q. Wang, Y.~Du, J.~Wang,
  \href{https://linkinghub.elsevier.com/retrieve/pii/S0360544220303042}{{LSTM}
  based long-term energy consumption prediction with periodicity}, Energy 197
  (2020-04) 117197.
\newblock \href {https://doi.org/10.1016/j.energy.2020.117197}
  {\path{doi:10.1016/j.energy.2020.117197}}.
\newline\urlprefix\url{https://linkinghub.elsevier.com/retrieve/pii/S0360544220303042}

\bibitem{salinas_deepar_2020}
D.~Salinas, V.~Flunkert, J.~Gasthaus, T.~Januschowski,
  \href{https://linkinghub.elsevier.com/retrieve/pii/S0169207019301888}{{DeepAR}:
  Probabilistic forecasting with autoregressive recurrent networks},
  International Journal of Forecasting 36~(3) (2020-07) 1181--1191.
\newblock \href {https://doi.org/10.1016/j.ijforecast.2019.07.001}
  {\path{doi:10.1016/j.ijforecast.2019.07.001}}.
\newline\urlprefix\url{https://linkinghub.elsevier.com/retrieve/pii/S0169207019301888}

\bibitem{herrero_2022}
J.~M. Santos-Herrero, J.~M. Lopez-Guede, I.~Flores~Abascal, E.~Zulueta,
  \href{https://www.nature.com/articles/s41598-022-12924-9}{Energy and thermal
  modelling of an office building to develop an artificial neural networks
  model}, Sci Rep 12~(1) (2022-05-27) 8935, publisher: Nature Publishing Group.
\newblock \href {https://doi.org/10.1038/s41598-022-12924-9}
  {\path{doi:10.1038/s41598-022-12924-9}}.
\newline\urlprefix\url{https://www.nature.com/articles/s41598-022-12924-9}

\bibitem{aryandoust_2022}
A.~Aryandoust, A.~Patt, S.~Pfenninger,
  \href{https://www.nature.com/articles/s42256-022-00552-x}{Enhanced
  spatio-temporal electric load forecasts using less data with active deep
  learning}, Nat Mach Intell 4~(11) (2022-11) 977--991, publisher: Nature
  Publishing Group.
\newblock \href {https://doi.org/10.1038/s42256-022-00552-x}
  {\path{doi:10.1038/s42256-022-00552-x}}.
\newline\urlprefix\url{https://www.nature.com/articles/s42256-022-00552-x}

\bibitem{monitor_germany_2023}
Monitoringbericht 2023 monitoringbericht gemäß § 63 abs. 3 i. v. m. § 35
  {EnWG} und § 48 abs. 3 i. v. m. § 53 abs. 3 {GWB} (2023-11-29).

\bibitem{GDPR_2016}
\href{http://data.europa.eu/eli/reg/2016/679/oj}{Regulation ({EU}) 2016/679 of
  the european parliament and of the council of 27 april 2016 on the protection
  of natural persons with regard to the processing of personal data and on the
  free movement of such data, and repealing directive 95/46/{EC} (general data
  protection regulation) (text with {EEA} relevance)} (2016-05-04).
\newline\urlprefix\url{http://data.europa.eu/eli/reg/2016/679/oj}

\bibitem{kezunovic_2013}
M.~Kezunovic, L.~Xie, S.~Grijalva,
  \href{https://ieeexplore.ieee.org/document/6629368}{The role of big data in
  improving power system operation and protection}, in: 2013 {IREP} Symposium
  Bulk Power System Dynamics and Control - {IX} Optimization, Security and
  Control of the Emerging Power Grid, 2013-08, pp. 1--9.
\newblock \href {https://doi.org/10.1109/IREP.2013.6629368}
  {\path{doi:10.1109/IREP.2013.6629368}}.
\newline\urlprefix\url{https://ieeexplore.ieee.org/document/6629368}

\bibitem{yu_2015}
N.~Yu, S.~Shah, R.~Johnson, R.~Sherick, M.~Hong, K.~Loparo,
  \href{https://ieeexplore.ieee.org/document/7131868}{Big data analytics in
  power distribution systems}, in: 2015 {IEEE} Power \& Energy Society
  Innovative Smart Grid Technologies Conference ({ISGT}), 2015-02, pp. 1--5.
\newblock \href {https://doi.org/10.1109/ISGT.2015.7131868}
  {\path{doi:10.1109/ISGT.2015.7131868}}.
\newline\urlprefix\url{https://ieeexplore.ieee.org/document/7131868}

\bibitem{albrecht_openmeter_2024}
M.~Albrecht, \href{https://www.openmeter.de/}{{openMeter} data platform}
  (2024).
\newline\urlprefix\url{https://www.openmeter.de/}

\bibitem{meier_reprasentative_1999}
H.~Meier, C.~Fünfgeld, T.~Adam, B.~Schieferdecker,
  \href{https://www.bdew.de/energie/standardlastprofile-strom/}{Repräsentative
  {VDEW}-lastprofile [representative {VDEW} load profiles]} (1999).
\newline\urlprefix\url{https://www.bdew.de/energie/standardlastprofile-strom/}

\bibitem{benitez2017classification}
I.~J. Ben{\'\i}tez~S{\'a}nchez, A.~Quijano~Lopez, I.~Delgado~Espinos, J.~L.
  Diez~Ruano, Classification of customers based on temporal load profile
  patterns, Cigre Science \& engineering 7 (2017) 143--148.

\bibitem{rasanen2009feature}
T.~R{\"a}s{\"a}nen, M.~Kolehmainen, Feature-based clustering for electricity
  use time series data, in: Adaptive and Natural Computing Algorithms: 9th
  International Conference, ICANNGA 2009, Kuopio, Finland, April 23-25, 2009,
  Revised Selected Papers 9, Springer, 2009, pp. 401--412.

\bibitem{yilmaz2019comparison}
S.~Yilmaz, J.~Chambers, M.~K. Patel, Comparison of clustering approaches for
  domestic electricity load profile characterisation-implications for demand
  side management, Energy 180 (2019) 665--677.

\bibitem{shi2020approach}
Y.~Shi, T.~Yu, Q.~Liu, H.~Zhu, F.~Li, Y.~Wu, An approach of electrical load
  profile analysis based on time series data mining, IEEE Access 8 (2020)
  209915--209925.

\bibitem{silipo2013big}
R.~Silipo, P.~Winters, Big data, smart energy, and predictive analytics, Time
  Series Prediction of Smart Energy Data 1~(37) (2013).

\bibitem{riedl2024cigre}
L.~Riedl, M.~Braun, P.~Hehlert, Efficient identification of customer types in
  energy consumption data: Leveraging dimensionality reduction and k-means
  clustering methods, in: CIGRE, Paris Session 2024, Ref C6-11443-2024, 2024,
  pp. 1--19.

\bibitem{hong_review_2020}
T.~Hong, P.~Pinson, Y.~Wang, R.~Weron, D.~Yang, H.~Zareipour,
  \href{https://ieeexplore.ieee.org/document/9218967/}{Energy forecasting: A
  review and outlook}, {IEEE} Open J. Power Energy 7 (2020) 376--388.
\newblock \href {https://doi.org/10.1109/OAJPE.2020.3029979}
  {\path{doi:10.1109/OAJPE.2020.3029979}}.
\newline\urlprefix\url{https://ieeexplore.ieee.org/document/9218967/}

\bibitem{gretton2006kernel}
A.~Gretton, K.~Borgwardt, M.~Rasch, B.~Sch{\"o}lkopf, A.~Smola, A kernel method
  for the two-sample-problem, Advances in neural information processing systems
  19 (2006).

\bibitem{mcinnes1802umap}
L.~McInnes, J.~Healy, J.~Melville, Umap: Uniform manifold approximation and
  projection for dimension reduction. arxiv 2018, arXiv preprint
  arXiv:1802.03426 10 (1802).

\bibitem{nti_electricity_2020}
I.~K. Nti, M.~Teimeh, O.~Nyarko-Boateng, A.~F. Adekoya,
  \href{https://doi.org/10.1186/s43067-020-00021-8}{Electricity load
  forecasting: a systematic review}, Journal of Electrical Systems and Inf
  Technol 7~(1) (2020-09-09) 13.
\newblock \href {https://doi.org/10.1186/s43067-020-00021-8}
  {\path{doi:10.1186/s43067-020-00021-8}}.
\newline\urlprefix\url{https://doi.org/10.1186/s43067-020-00021-8}

\bibitem{box_2016}
G.~E.~P. Box, G.~M. Jenkins, G.~C. Reinsel, G.~M. Ljung, Time series analysis:
  forecasting and control, fifth edition Edition, Wiley series in probability
  and statistics, John Wiley \& Sons, Inc.

\bibitem{yunfan_li_2024}
Y.~Li,
  \href{https://iopscience.iop.org/article/10.1088/1742-6596/2711/1/012012}{Energy
  consumption forecasting with deep learning}, J. Phys.: Conf. Ser. 2711~(1)
  (2024-02-01) 012012.
\newblock \href {https://doi.org/10.1088/1742-6596/2711/1/012012}
  {\path{doi:10.1088/1742-6596/2711/1/012012}}.
\newline\urlprefix\url{https://iopscience.iop.org/article/10.1088/1742-6596/2711/1/012012}

\bibitem{de_oliveira_arima_2018}
E.~M. De~Oliveira, F.~L. Cyrino~Oliveira,
  \href{https://linkinghub.elsevier.com/retrieve/pii/S0360544217320820}{Forecasting
  mid-long term electric energy consumption through bagging {ARIMA} and
  exponential smoothing methods}, Energy 144 (2018-02) 776--788.
\newblock \href {https://doi.org/10.1016/j.energy.2017.12.049}
  {\path{doi:10.1016/j.energy.2017.12.049}}.
\newline\urlprefix\url{https://linkinghub.elsevier.com/retrieve/pii/S0360544217320820}

\bibitem{dong_2005}
B.~Dong, C.~Cao, S.~E. Lee,
  \href{https://linkinghub.elsevier.com/retrieve/pii/S0378778804002981}{Applying
  support vector machines to predict building energy consumption in tropical
  region}, Energy and Buildings 37~(5) (2005-05) 545--553.
\newblock \href {https://doi.org/10.1016/j.enbuild.2004.09.009}
  {\path{doi:10.1016/j.enbuild.2004.09.009}}.
\newline\urlprefix\url{https://linkinghub.elsevier.com/retrieve/pii/S0378778804002981}

\bibitem{jiao_2022}
J.~Jiao, H.~Brugger, M.~Behrisch, W.~Eichhammer, Identifying drivers of
  residential energy consumption by explainable energy demand forecasting,
  {ECEEE} {SUMMER} {STUDY} {PROCEEDINGS} (2022).

\bibitem{tso_yan_2007}
G.~K. Tso, K.~K. Yau,
  \href{https://linkinghub.elsevier.com/retrieve/pii/S0360544206003288}{Predicting
  electricity energy consumption: A comparison of regression analysis, decision
  tree and neural networks}, Energy 32~(9) (2007-09) 1761--1768.
\newblock \href {https://doi.org/10.1016/j.energy.2006.11.010}
  {\path{doi:10.1016/j.energy.2006.11.010}}.
\newline\urlprefix\url{https://linkinghub.elsevier.com/retrieve/pii/S0360544206003288}

\bibitem{li_building_2017}
C.~Li, Z.~Ding, D.~Zhao, J.~Yi, G.~Zhang,
  \href{https://www.mdpi.com/1996-1073/10/10/1525}{Building energy consumption
  prediction: An extreme deep learning approach}, Energies 10~(10) (2017-10-07)
  1525.
\newblock \href {https://doi.org/10.3390/en10101525}
  {\path{doi:10.3390/en10101525}}.
\newline\urlprefix\url{https://www.mdpi.com/1996-1073/10/10/1525}

\bibitem{souhe_2021}
F.~G.~Y. Souhe, C.~F. Mbey, A.~T. Boum, P.~Ele,
  \href{https://econjournals.com/index.php/ijeep/article/view/11761}{{FORECASTING}
  {OF} {ELECTRICAL} {ENERGY} {CONSUMPTION} {OF} {HOUSEHOLDS} {IN} a {SMART}
  {GRID}}, {IJEEP} 11~(6) (2021-11-05) 221--233.
\newblock \href {https://doi.org/10.32479/ijeep.11761}
  {\path{doi:10.32479/ijeep.11761}}.
\newline\urlprefix\url{https://econjournals.com/index.php/ijeep/article/view/11761}

\bibitem{singh_2018}
S.~Singh, A.~Yassine, \href{https://www.mdpi.com/1996-1073/11/2/452}{Big data
  mining of energy time series for behavioral analytics and energy consumption
  forecasting}, Energies 11~(2) (2018-02-20) 452.
\newblock \href {https://doi.org/10.3390/en11020452}
  {\path{doi:10.3390/en11020452}}.
\newline\urlprefix\url{https://www.mdpi.com/1996-1073/11/2/452}

\bibitem{muralitharan_2018}
K.~Muralitharan, R.~Sakthivel, R.~Vishnuvarthan,
  \href{https://linkinghub.elsevier.com/retrieve/pii/S0925231217313681}{Neural
  network based optimization approach for energy demand prediction in smart
  grid}, Neurocomputing 273 (2018-01) 199--208.
\newblock \href {https://doi.org/10.1016/j.neucom.2017.08.017}
  {\path{doi:10.1016/j.neucom.2017.08.017}}.
\newline\urlprefix\url{https://linkinghub.elsevier.com/retrieve/pii/S0925231217313681}

\bibitem{rahman_2018}
A.~Rahman, V.~Srikumar, A.~D. Smith,
  \href{https://linkinghub.elsevier.com/retrieve/pii/S0306261917317658}{Predicting
  electricity consumption for commercial and residential buildings using deep
  recurrent neural networks}, Applied Energy 212 (2018-02) 372--385.
\newblock \href {https://doi.org/10.1016/j.apenergy.2017.12.051}
  {\path{doi:10.1016/j.apenergy.2017.12.051}}.
\newline\urlprefix\url{https://linkinghub.elsevier.com/retrieve/pii/S0306261917317658}

\bibitem{vaswani_attention_2017}
A.~Vaswani, N.~Shazeer, N.~Parmar, J.~Uszkoreit, L.~Jones, A.~N. Gomez,
  L.~Kaiser, I.~Polosukhin, \href{http://arxiv.org/abs/1706.03762}{Attention is
  all you need} (2017-12-05).
\newblock \href {http://arxiv.org/abs/1706.03762 [cs]} {\path{arXiv:1706.03762
  [cs]}}.
\newline\urlprefix\url{http://arxiv.org/abs/1706.03762}

\bibitem{li_transformer_2019}
S.~Li, X.~Jin, Y.~Xuan, X.~Zhou, W.~Chen, Y.-X. Wang, X.~Yan, Enhancing the
  locality and breaking the memory bottleneck of transformer on time series
  forecasting, 33rd Conference on Neural Information Processing Systems
  {NeurIPS} 2019 (2019).

\bibitem{qureshi_LSTM_2024}
M.~Qureshi, M.~A. Arbab, S.~u. Rehman,
  \href{https://www.nature.com/articles/s41598-024-56602-4}{Deep learning-based
  forecasting of electricity consumption}, Sci Rep 14~(1) (2024-03-18) 6489,
  publisher: Nature Publishing Group.
\newblock \href {https://doi.org/10.1038/s41598-024-56602-4}
  {\path{doi:10.1038/s41598-024-56602-4}}.
\newline\urlprefix\url{https://www.nature.com/articles/s41598-024-56602-4}

\bibitem{lai_liu_2018}
G.~Lai, W.-C. Chang, Y.~Yang, H.~Liu,
  \href{https://dl.acm.org/doi/10.1145/3209978.3210006}{Modeling long- and
  short-term temporal patterns with deep neural networks}, in: The 41st
  International {ACM} {SIGIR} Conference on Research \& Development in
  Information Retrieval, {SIGIR} '18, Association for Computing Machinery,
  2018-06-27, pp. 95--104.
\newblock \href {https://doi.org/10.1145/3209978.3210006}
  {\path{doi:10.1145/3209978.3210006}}.
\newline\urlprefix\url{https://dl.acm.org/doi/10.1145/3209978.3210006}

\bibitem{ramos_2023}
P.~V.~B. Ramos, S.~M. Villela, W.~N. Silva, B.~H. Dias,
  \href{https://linkinghub.elsevier.com/retrieve/pii/S0306261923010693}{Residential
  energy consumption forecasting using deep learning models}, Applied Energy
  350 (2023-11) 121705.
\newblock \href {https://doi.org/10.1016/j.apenergy.2023.121705}
  {\path{doi:10.1016/j.apenergy.2023.121705}}.
\newline\urlprefix\url{https://linkinghub.elsevier.com/retrieve/pii/S0306261923010693}

\bibitem{ou_ali_autoencoder_2024}
I.~H. Ou~Ali, A.~Agga, M.~Ouassaid, M.~Maaroufi, A.~Elrashidi, H.~Kotb,
  \href{https://www.frontiersin.org/articles/10.3389/fenrg.2024.1323357/full}{Predicting
  short-term energy usage in a smart home using hybrid deep learning models},
  Front. Energy Res. 12 (2024-09-05) 1323357.
\newblock \href {https://doi.org/10.3389/fenrg.2024.1323357}
  {\path{doi:10.3389/fenrg.2024.1323357}}.
\newline\urlprefix\url{https://www.frontiersin.org/articles/10.3389/fenrg.2024.1323357/full}

\bibitem{wang_lstm_informer_2023}
K.~Wang, J.~Zhang, X.~Li, Y.~Zhang,
  \href{https://www.mdpi.com/2079-9292/12/10/2175}{Long-term power load
  forecasting using {LSTM}-informer with ensemble learning}, Electronics
  12~(10) (2023-01) 2175, number: 10 Publisher = Multidisciplinary Digital
  Publishing Institute.
\newblock \href {https://doi.org/10.3390/electronics12102175}
  {\path{doi:10.3390/electronics12102175}}.
\newline\urlprefix\url{https://www.mdpi.com/2079-9292/12/10/2175}

\bibitem{wu_autoformer_2021}
H.~Wu, J.~Xu, J.~Wang, M.~Long, Autoformer: Decomposition transformers with
  auto-correlation for long-term series forecasting, 35th Conference on Neural
  Information Processing Systems (2021).

\bibitem{zhou_informer_2021}
H.~Zhou, S.~Zhang, J.~Peng, S.~Zhang, J.~Li, H.~Xiong, W.~Zhang,
  \href{https://ojs.aaai.org/index.php/AAAI/article/view/17325}{Informer:
  Beyond efficient transformer for long sequence time-series forecasting},
  Proceedings of the {AAAI} Conference on Artificial Intelligence 35~(12)
  (2021-05-18) 11106--11115, number: 12.
\newblock \href {https://doi.org/10.1609/aaai.v35i12.17325}
  {\path{doi:10.1609/aaai.v35i12.17325}}.
\newline\urlprefix\url{https://ojs.aaai.org/index.php/AAAI/article/view/17325}

\bibitem{arjovsky2017wasserstein}
M.~Arjovsky, S.~Chintala, L.~Bottou, Wasserstein generative adversarial
  networks, in: International conference on machine learning, PMLR, 2017, pp.
  214--223.

\bibitem{gulrajani2017improved}
I.~Gulrajani, F.~Ahmed, M.~Arjovsky, V.~Dumoulin, A.~C. Courville, Improved
  training of wasserstein gans, Advances in neural information processing
  systems 30 (2017).

\bibitem{hochreiter1997long}
S.~Hochreiter, J.~Schmidhuber, Long short-term memory, Neural computation 9~(8)
  (1997) 1735--1780.

\bibitem{HoJainAbbeel2020}
J.~Ho, A.~Jain, P.~Abbeel, \href{https://arxiv.org/abs/2006.11239}{Denoising
  diffusion probabilistic models}, CoRR abs/2006.11239 (2020).
\newblock \href {http://arxiv.org/abs/2006.11239} {\path{arXiv:2006.11239}}.
\newline\urlprefix\url{https://arxiv.org/abs/2006.11239}

\bibitem{NicholDhariwall2021}
A.~Nichol, P.~Dhariwal, \href{https://arxiv.org/abs/2102.09672}{Improved
  denoising diffusion probabilistic models}, CoRR abs/2102.09672 (2021).
\newblock \href {http://arxiv.org/abs/2102.09672} {\path{arXiv:2102.09672}}.
\newline\urlprefix\url{https://arxiv.org/abs/2102.09672}

\bibitem{geurten_hmm_2010}
B.~R.~H. Geurten, R.~Kern, E.~Braun, M.~Egelhaaf, A syntax of hoverfly flight
  prototypes, {THE} {JOURNAL} {OF} {EXPERIMENTAL} {BIOLOGY} 213 (2010)
  2461--247515.
\newblock \href {https://doi.org/doi:10.1242/jeb.036079}
  {\path{doi:doi:10.1242/jeb.036079}}.

\bibitem{braun_hmm_2010}
E.~Braun, B.~Geurten, M.~Egelhaaf,
  \href{https://dx.plos.org/10.1371/journal.pone.0009361}{Identifying
  prototypical components in behaviour using clustering algorithms}, {PLoS}
  {ONE} 5~(2) (2010-02-22) e9361.
\newblock \href {https://doi.org/10.1371/journal.pone.0009361}
  {\path{doi:10.1371/journal.pone.0009361}}.
\newline\urlprefix\url{https://dx.plos.org/10.1371/journal.pone.0009361}

\bibitem{gabrielski_markov_2023}
J.~Gabrielski, U.~Häger,
  \href{https://ieeexplore.ieee.org/document/10294910}{A markov chain model for
  imputation of electricity consumption time series}, in: 2023 58th
  International Universities Power Engineering Conference ({UPEC}), 2023-08,
  pp. 1--6.
\newblock \href {https://doi.org/10.1109/UPEC57427.2023.10294910}
  {\path{doi:10.1109/UPEC57427.2023.10294910}}.
\newline\urlprefix\url{https://ieeexplore.ieee.org/document/10294910}

\bibitem{Hothorn2014}
T.~Hothorn, T.~Kneib, P.~Bühlmann, Conditional {{Transformation Models}},
  Journal of the Royal Statistical Society Series B: Statistical Methodology
  76~(1) (2014-01) 3--27.
\newblock \href {https://doi.org/10.1111/rssb.12017}
  {\path{doi:10.1111/rssb.12017}}.

\bibitem{Hothorn2018}
T.~Hothorn, L.~Möst, P.~Bühlmann, Most {{Likely Transformations}},
  Scandinavian Journal of Statistics 45~(1) (2018) 110--134.
\newblock \href {https://doi.org/10.1111/sjos.12291}
  {\path{doi:10.1111/sjos.12291}}.

\bibitem{Papamakarios2018}
G.~Papamakarios, T.~Pavlakou, I.~Murray, Masked {{Autoregressive Flow}} for
  {{Density Estimation}} (2018-06-14).
\newblock \href {http://arxiv.org/abs/1705.07057} {\path{arXiv:1705.07057}}.

\bibitem{Papamakarios2021}
G.~Papamakarios, E.~Nalisnick, D.~J. Rezende, S.~Mohamed, B.~Lakshminarayanan,
  \href{http://jmlr.org/papers/v22/19-1028.html}{Normalizing {{Flows}} for
  {{Probabilistic Modeling}} and {{Inference}}}, Journal of Machine Learning
  Research 22~(57) (2021) 1--64.
\newline\urlprefix\url{http://jmlr.org/papers/v22/19-1028.html}

\bibitem{Farouki2012}
R.~T. Farouki, The {{Bernstein Polynomial Basis}}: {{A Centennial
  Retrospective}}, Computer Aided Geometric Design 29~(6) (2012-08) 379--419.
\newblock \href {https://doi.org/10.1016/j.cagd.2012.03.001}
  {\path{doi:10.1016/j.cagd.2012.03.001}}.

\bibitem{fischer_synpro_2020}
D.~Fischer, A.~Surmann, W.~Biener, O.~Selinger-Lutz,
  \href{https://www.sciencedirect.com/science/article/pii/S0378778819335777}{From
  residential electric load profiles to flexibility profiles – a stochastic
  bottom-up approach}, Energy and Buildings 224 (2020-10-01) 110133.
\newblock \href {https://doi.org/10.1016/j.enbuild.2020.110133}
  {\path{doi:10.1016/j.enbuild.2020.110133}}.
\newline\urlprefix\url{https://www.sciencedirect.com/science/article/pii/S0378778819335777}

\bibitem{sylaski_datarepo_2024}
P.~Hehlert, B.~Gerhards, N.~Popkov,
  \href{https://doi.org/10.25625/A5VZA9}{{Sylas KI synthetic powerprofiles
  OpenMeter}} (2024).
\newblock \href {https://doi.org/10.25625/A5VZA9} {\path{doi:10.25625/A5VZA9}}.
\newline\urlprefix\url{https://doi.org/10.25625/A5VZA9}

\bibitem{li_review_federatedlearning_2020}
L.~Li, Y.~Fan, M.~Tse, K.-Y. Lin,
  \href{https://www.sciencedirect.com/science/article/pii/S0360835220305532}{A
  review of applications in federated learning}, Computers \& Industrial
  Engineering 149 (2020-11-01) 106854.
\newblock \href {https://doi.org/10.1016/j.cie.2020.106854}
  {\path{doi:10.1016/j.cie.2020.106854}}.
\newline\urlprefix\url{https://www.sciencedirect.com/science/article/pii/S0360835220305532}

\bibitem{mammen_federated_2021}
P.~M. Mammen, \href{http://arxiv.org/abs/2101.05428}{Federated learning:
  Opportunities and challenges} (2021-01-14).
\newblock \href {http://arxiv.org/abs/2101.05428 [cs]} {\path{arXiv:2101.05428
  [cs]}}, \href {https://doi.org/10.48550/arXiv.2101.05428}
  {\path{doi:10.48550/arXiv.2101.05428}}.
\newline\urlprefix\url{http://arxiv.org/abs/2101.05428}

\bibitem{manning_introduction_2008}
C.~D. Manning, P.~Raghavan, H.~Schütze, Introduction to information retrieval,
  Cambridge university press, 2008.

\bibitem{guo2016entity}
C.~Guo, F.~Berkhahn, Entity embeddings of categorical variables, arXiv preprint
  arXiv:1604.06737 (2016).

\bibitem{goodfellow2020generative}
I.~Goodfellow, J.~Pouget-Abadie, M.~Mirza, B.~Xu, D.~Warde-Farley, S.~Ozair,
  A.~Courville, Y.~Bengio, Generative adversarial networks, Communications of
  the ACM 63~(11) (2020) 139--144.

\bibitem{zhu2017unpaired}
J.-Y. Zhu, T.~Park, P.~Isola, A.~A. Efros, Unpaired image-to-image translation
  using cycle-consistent adversarial networks, in: Proceedings of the IEEE
  international conference on computer vision, 2017, pp. 2223--2232.

\bibitem{lin2020relevant}
E.~Lin, C.-H. Lin, H.-Y. Lane, Relevant applications of generative adversarial
  networks in drug design and discovery: molecular de novo design,
  dimensionality reduction, and de novo peptide and protein design, Molecules
  25~(14) (2020) 3250.

\bibitem{villani2008optimal}
C.~Villani, et~al., Optimal transport: old and new, Vol. 338, Springer, 2008.

\bibitem{ronneberger2015u}
O.~Ronneberger, P.~Fischer, T.~Brox, U-net: Convolutional networks for
  biomedical image segmentation, in: Medical image computing and
  computer-assisted intervention--MICCAI 2015: 18th international conference,
  Munich, Germany, October 5-9, 2015, proceedings, part III 18, Springer, 2015,
  pp. 234--241.

\bibitem{Kingma2018}
D.~P. Kingma, P.~Dhariwal, Glow: {{Generative Flow}} with {{Invertible}} 1x1
  {{Convolutions}} (2018-07-10).
\newblock \href {http://arxiv.org/abs/1807.03039} {\path{arXiv:1807.03039}},
  \href {https://doi.org/10.48550/arXiv.1807.03039}
  {\path{doi:10.48550/arXiv.1807.03039}}.

\bibitem{Oord2016}
A.~Van Den~Oord, S.~Dieleman, H.~Zen, K.~Simonyan, O.~Vinyals, A.~Graves,
  N.~Kalchbrenner, A.~Senior, K.~Kavukcuoglu, et~al., Wavenet: A generative
  model for raw audio, arXiv preprint arXiv:1609.03499 12 (2016).

\bibitem{Rezende2016}
D.~J. Rezende, S.~Mohamed, Variational {{Inference}} with {{Normalizing Flows}}
  (2016-06-14).
\newblock \href {http://arxiv.org/abs/1505.05770} {\path{arXiv:1505.05770}}.

\bibitem{Kobyzev2021}
I.~Kobyzev, S.~J. Prince, M.~A. Brubaker, Normalizing {{Flows}}: {{An
  Introduction}} and {{Review}} of {{Current Methods}}, IEEE Transactions on
  Pattern Analysis and Machine Intelligence 43~(11) (2021-11) 3964--3979.
\newblock \href {http://arxiv.org/abs/1908.09257} {\path{arXiv:1908.09257}},
  \href {https://doi.org/10.1109/TPAMI.2020.2992934}
  {\path{doi:10.1109/TPAMI.2020.2992934}}.

\bibitem{Kneib2021}
T.~Kneib, A.~Silbersdorff, B.~Säfken, Rage {{Against}} the {{Mean}} – {{A
  Review}} of {{Distributional Regression Approaches}}, Econometrics and
  Statistics (2021-08-10).
\newblock \href {https://doi.org/10.1016/j.ecosta.2021.07.006}
  {\path{doi:10.1016/j.ecosta.2021.07.006}}.

\bibitem{Baumann2021}
P.~F. Baumann, T.~Hothorn, D.~R{\"u}gamer, Deep conditional transformation
  models, in: Joint European Conference on Machine Learning and Knowledge
  Discovery in Databases, Springer, 2021, pp. 3--18.

\bibitem{Sick2021}
B.~Sick, T.~Hothorn, O.~Dürr, Deep {{Transformation Models}}: {{Tackling
  Complex Regression Problems}} with {{Neural Network Based Transformation
  Models}}, in: 2020 25th {{International Conference}} on {{Pattern
  Recognition}} ({{ICPR}}), IEEE, IEEE, 2021-01, pp. 2476--2481.
\newblock \href {https://doi.org/10/gpd2v4} {\path{doi:10/gpd2v4}}.

\bibitem{Arpogaus2023}
M.~Arpogaus, M.~Voss, B.~Sick, M.~Nigge-Uricher, O.~Dürr, Short-{{Term Density
  Forecasting}} of {{Low-Voltage Load Using Bernstein-Polynomial Normalizing
  Flows}}, IEEE Transactions on Smart Grid 14~(6) (2023-11) 4902--4911.
\newblock \href {http://arxiv.org/abs/2204.13939} {\path{arXiv:2204.13939}},
  \href {https://doi.org/10.1109/tsg.2023.3254890}
  {\path{doi:10.1109/tsg.2023.3254890}}.

\bibitem{Kook2021}
L.~Kook, L.~Herzog, T.~Hothorn, O.~Dürr, B.~Sick, Deep and interpretable
  regression models for ordinal outcomes (2021-04-20).
\newblock \href {http://arxiv.org/abs/2010.08376} {\path{arXiv:2010.08376}},
  \href {https://doi.org/10.48550/arXiv.2010.08376}
  {\path{doi:10.48550/arXiv.2010.08376}}.

\bibitem{Rugamer2023}
D.~Rugamer, P.~F.~M. Baumann, T.~Kneib, T.~Hothorn, Probabilistic time series
  forecasts with autoregressive transformation models, Statistics and Computing
  33~(2) (2023-02-04) 37.
\newblock \href {https://doi.org/10.1007/s11222-023-10212-8}
  {\path{doi:10.1007/s11222-023-10212-8}}.

\bibitem{Kingma2017a}
D.~P. Kingma, J.~Ba, Adam: {{A Method}} for {{Stochastic Optimization}}
  (2017-01-29).
\newblock \href {http://arxiv.org/abs/1412.6980} {\path{arXiv:1412.6980}}.

\bibitem{Germain2015}
M.~Germain, K.~Gregor, I.~Murray, H.~Larochelle, {{MADE}}: {{Masked
  Autoencoder}} for {{Distribution Estimation}} (2015-06-05).
\newblock \href {http://arxiv.org/abs/1502.03509} {\path{arXiv:1502.03509}}.

\bibitem{Sick2025}
B.~Sick, O.~Dürr, Interpretable {{Neural Causal Models}} with {{TRAM-DAGs}}
  (2025-03-20).
\newblock \href {http://arxiv.org/abs/2503.16206} {\path{arXiv:2503.16206}},
  \href {https://doi.org/10.48550/arXiv.2503.16206}
  {\path{doi:10.48550/arXiv.2503.16206}}.

\bibitem{schreiber2018pomegranate}
J.~Schreiber, Pomegranate: fast and flexible probabilistic modeling in python,
  Journal of Machine Learning Research 18~(164) (2018) 1--6.

\end{thebibliography}

\newpage
\section{Supplementary Figures}
\setcounter{figure}{0}
\renewcommand{\thefigure}{S\arabic{figure}}

\begin{figure}[!ht]
\centering
\resizebox{0.65\textwidth}{!}{\includegraphics{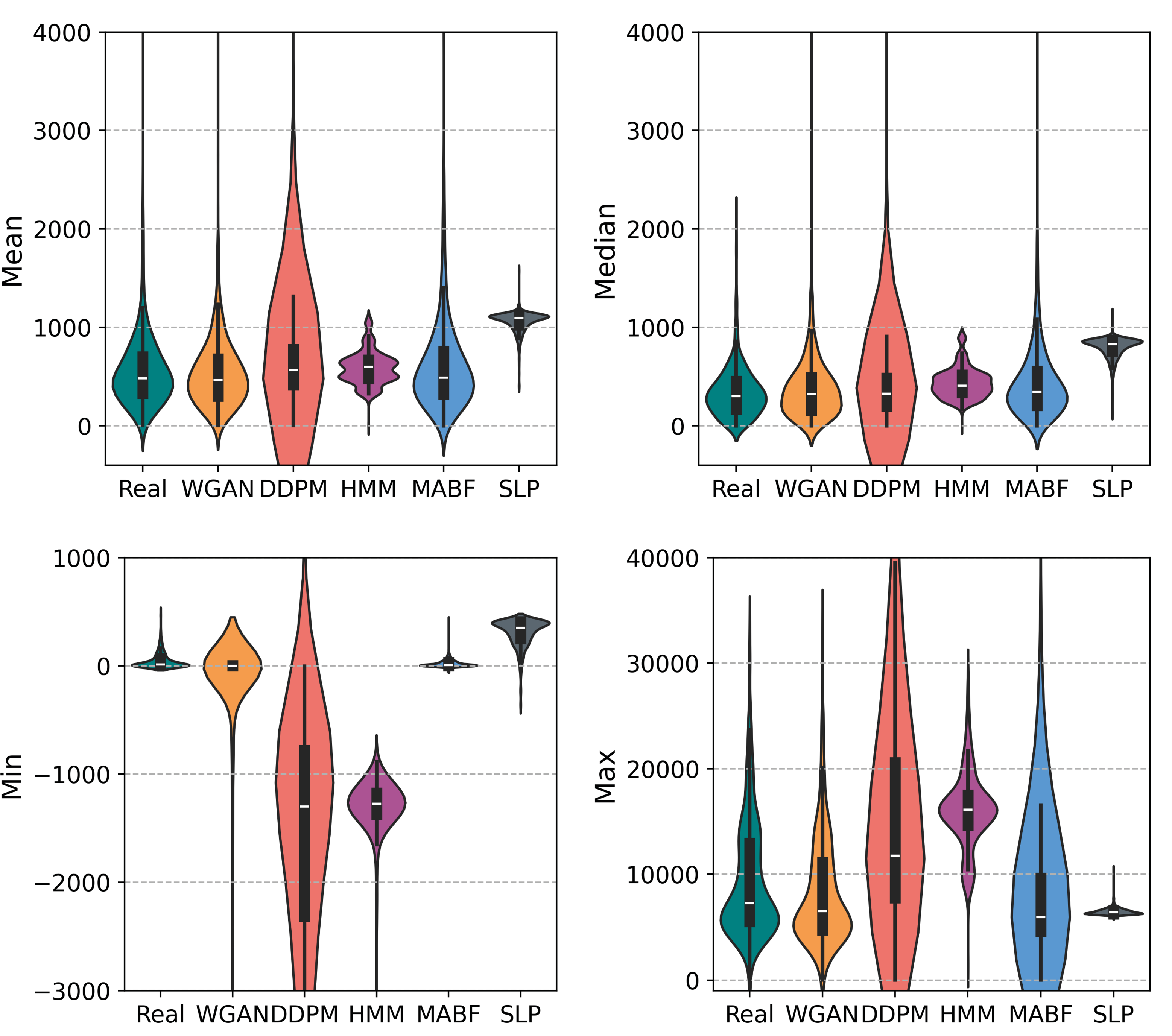}}
\caption{Simple statistical metrics -Violin-plots for simple statistical analysis (see also Table \ref{tab:simple_stats}) of power consumption values for each consumer from the respective ML model compared to the \gls{om} data.}
\label{fig:easy_stats}
\end{figure}

\begin{table}[ht!]
\centering
\small
\begin{tabular}{ccccccc}
\hline
Metric&OM&WGAN&DDPM&HMM&MABF&SLP\\\

$\pm$SD\\
\hline
sig.&&\textit{n.s.}&***&***&\textit{n.s.}&***\\
Mean &  0.57$\pm$0.46&0.55$\pm$0.44&0.76$\pm$2.67&0.59$\pm$0.17&0.63$\pm$ 0.54&1.06$\pm$0.10\\
Median&0.35$\pm$0.27&0.38$\pm$0.36& 0.47$\pm$2.14&0.44$\pm$0.15&0.44$\pm$0.42&0.79$\pm$0.11\\
Min&0.05$\pm$0.07&-0.17$\pm$0.68&-1.81$\pm$2.28&-1.33$\pm$0.43& 0.02$\pm$0.03&0.31$\pm$0.12\\
Max&9.5$\pm$5.56&8.40$\pm$5.18&16.3$\pm$28.9&16.1$\pm$3.6& 10.2$\pm$25.3&6.50$\pm$0.34\\

\hline
\end{tabular}
\caption{Simple statistics on average power consumption values from average consumers for each dataset. Averaged means between \gls{om} and \gls{WGAN} or \gls{MABF} were not significantly different. Consumers generated by \gls{DDPM}, \gls{HMM} or \gls{SLP} exhibited highly significant increase in mean power consumption. Distributions of means from each model were compared to the \gls{om} data; Mann-Whitney-U-test with Bonferroni correction: P$>$0.05 = non significant (\textit{n.s.}), P$<$0.001 = ***)}
\label{tab:simple_stats}
\end{table}\textbf{}

\begin{figure}[!ht]
\centering
\resizebox{0.85\textwidth}{!}{\includegraphics{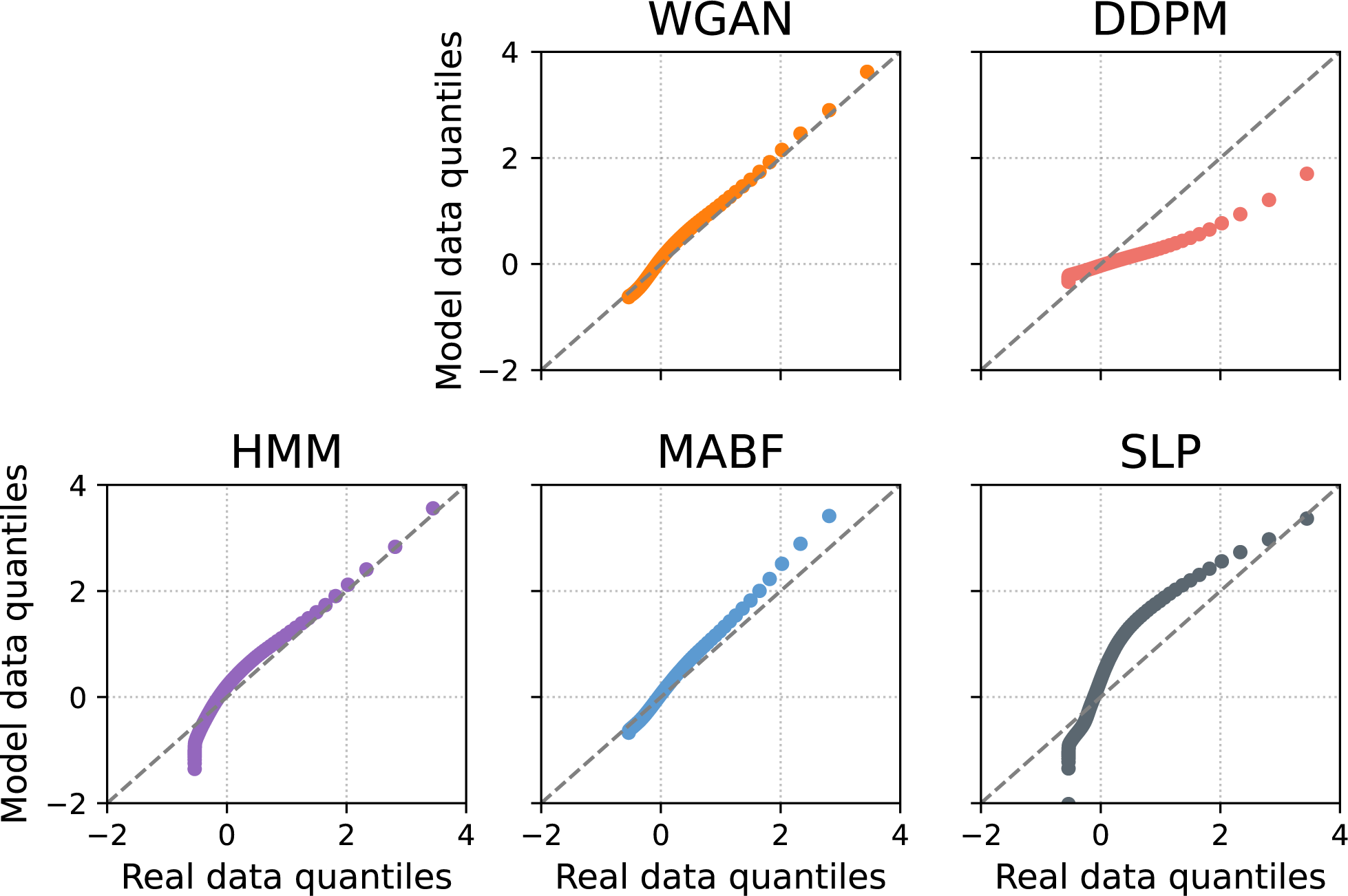}}
\caption{Q-Q-Plots of power consumption - Pooled normalized data from the \gls{om} dataset and their respective time-series surrogates (from \gls{WGAN}, \gls{DDPM}, \gls{HMM}, \gls{MABF}, and \gls{SLP}.)}
\label{fig:qq_plot}
\end{figure}

\begin{figure}[!ht]
\centering
\resizebox{0.70\textwidth}{!}{\includegraphics{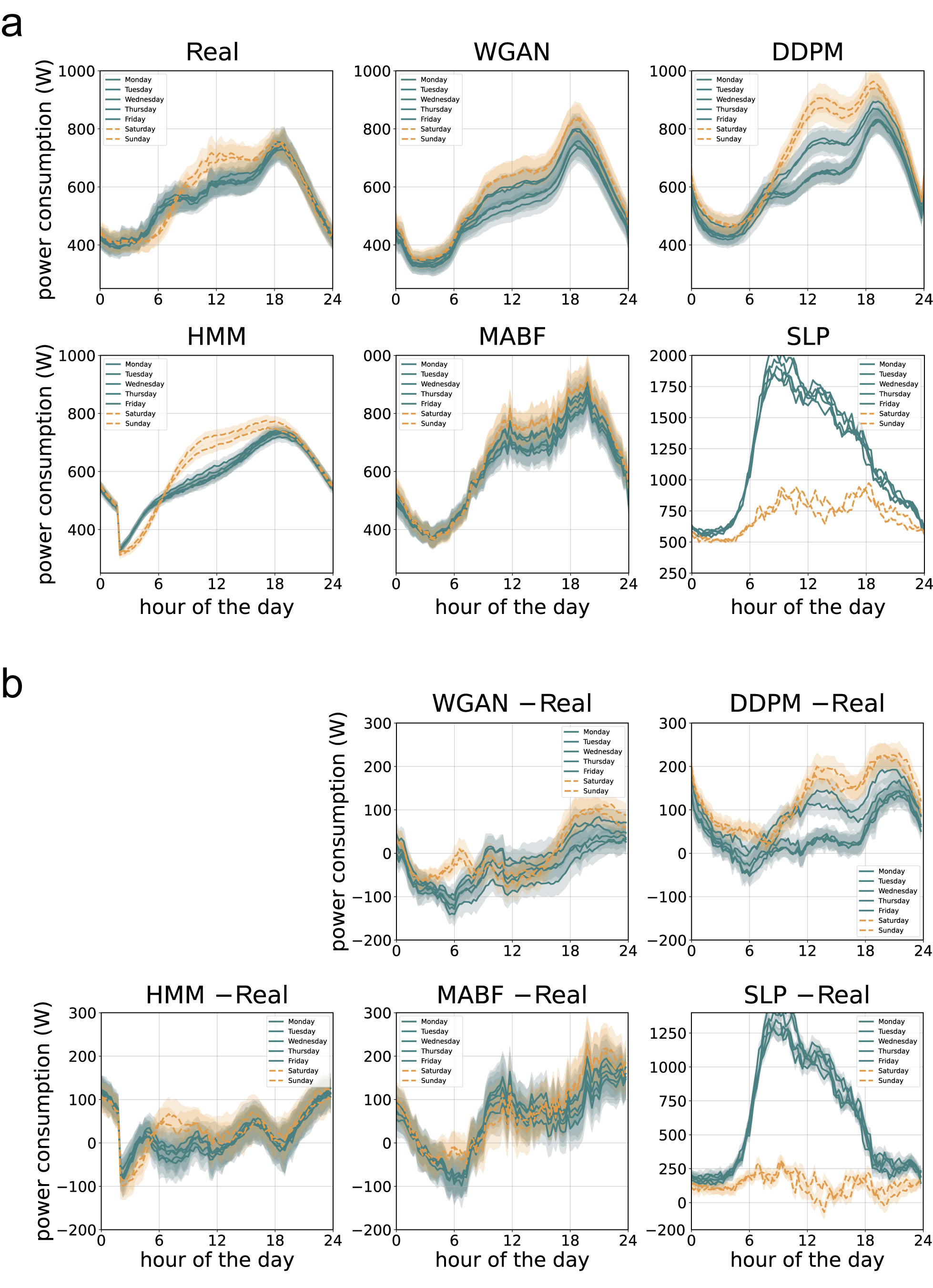}}
\caption{Average daily power consumption. a) Line plot of average power consumption (W) per hour of the day (+95\% confidence interval) for each day of the week and respective ML model. Average is calculated for all consumer profiles. b) Differential line plot to visualize over-/under-power consumption for each time of the day for the respective ML model.}
\label{fig:avg_power_diff}
\end{figure}

\begin{figure}[!ht]
\centering
\resizebox{1.0 \textwidth}{!}{\includegraphics{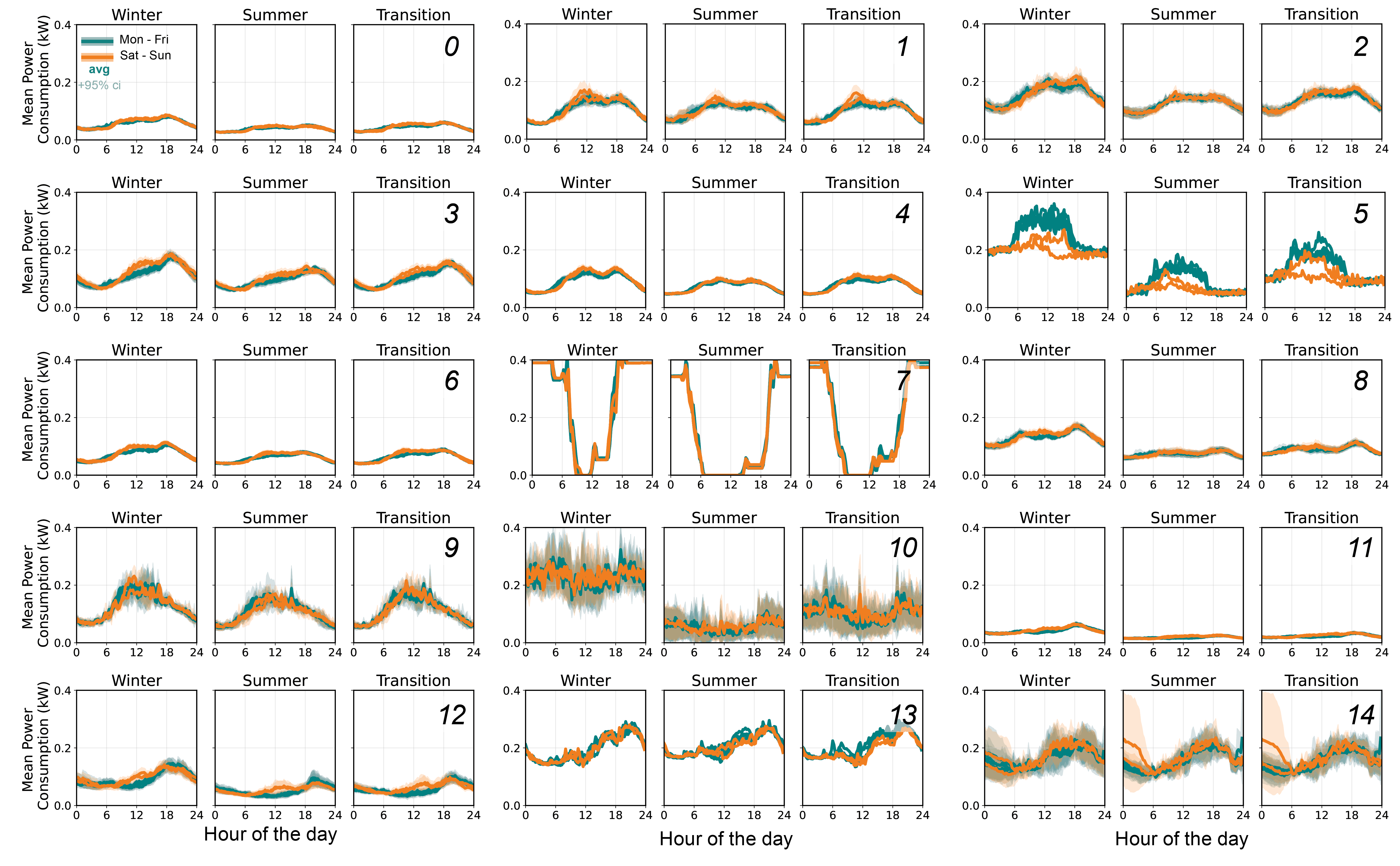}}
\caption{Consumer types OM dataset - Shown are averages +95\% CI from typical weeks from clustered households (PCA with 5 PCs, k=15, k-means+).}
\label{fig:clust_OM}
\end{figure}

\begin{figure}[!ht]
\centering
\resizebox{0.99\textwidth}{!}{\includegraphics{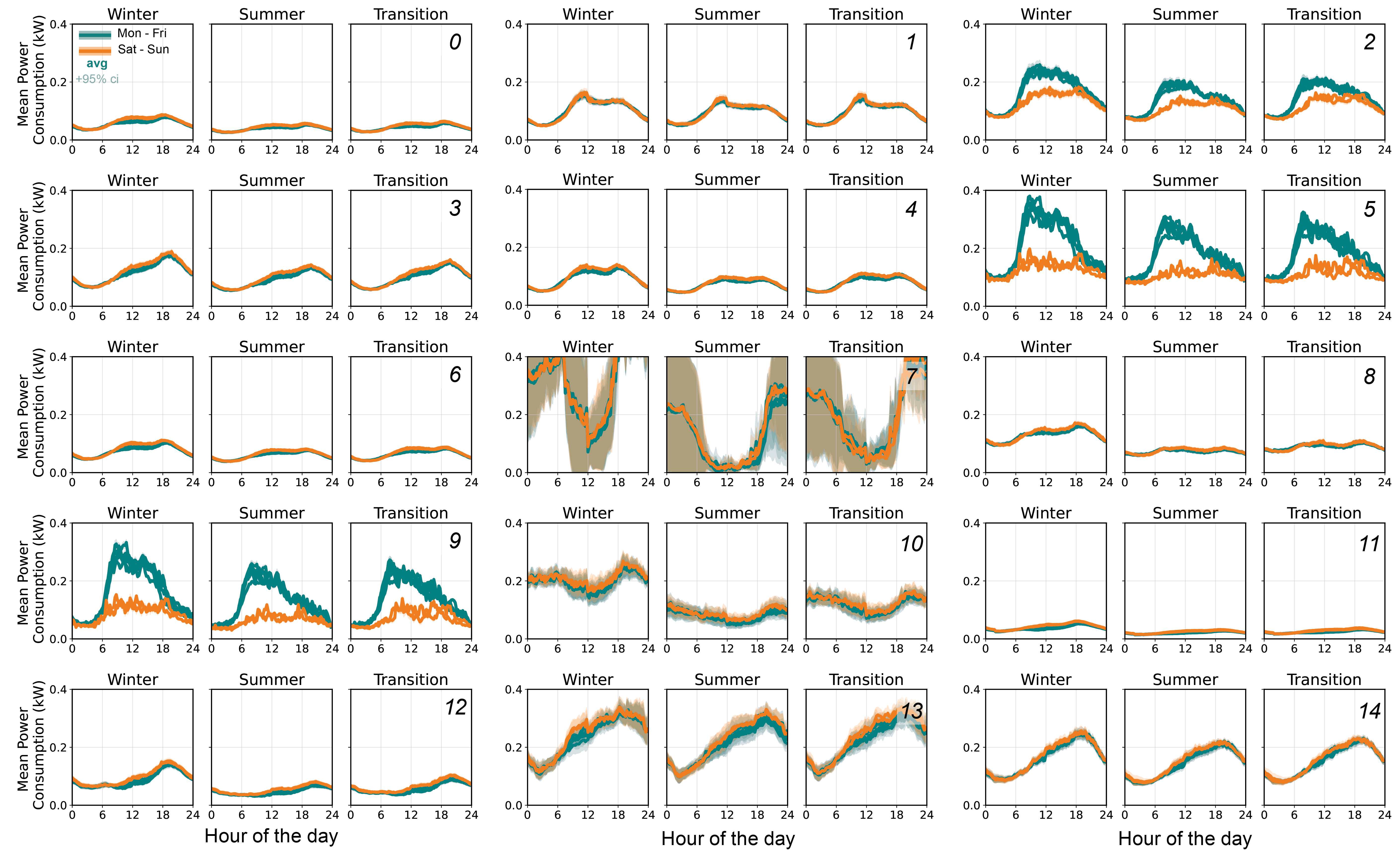}}
\caption{Consumer typing - Classification of power consumers (only \gls{om} data used in clustering, PCA with 5 PCs, k=15, k-means+). Synthetic profiles were assigned to OM centroids. Shown are average typical weeks (+ 95\% CI) from all (\gls{om} and all tested models) power profiles.}
\label{fig:clust_all_on_OM}
\end{figure}

\begin{figure}[!ht]
\centering
\resizebox{1.0\textwidth}{!}{\includegraphics{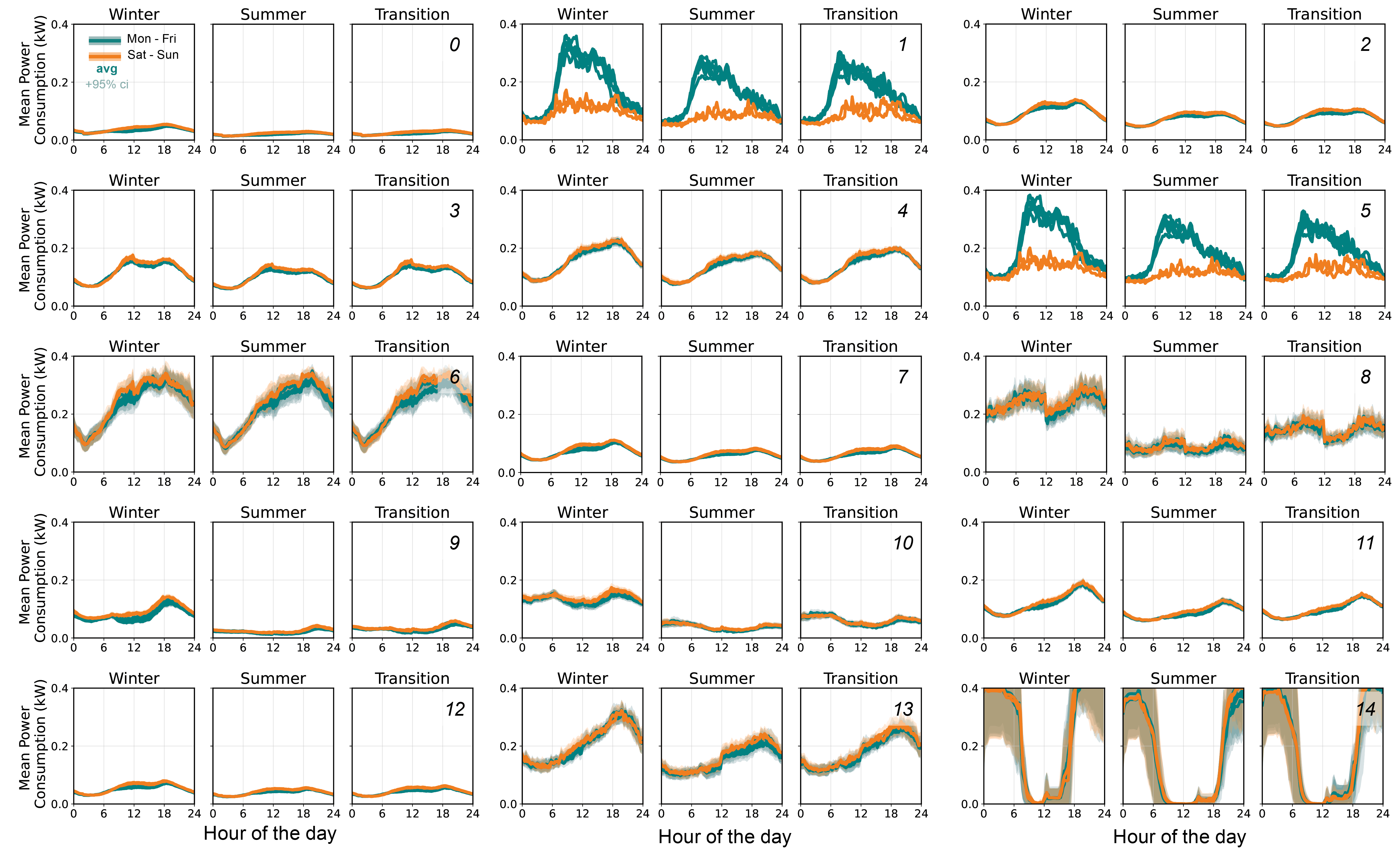}}
\caption{Mixed consumer typing - Shown are averages $\pm$95\% confidence interval from typical weeks of the respective consumer types (clustering with all data (OM + synthetic data), PCA with 5 PCs, k=15, k-means+).}
\label{fig:clust_all_blended}
\end{figure}

\begin{table}[ht]
    \centering
    \begin{tabular}{lccccc}
        \hline
       & WGAN & DDPM & HMM & MABF & SLP \\
        \hline
        MAE & 489.6$\pm$457.4 & 734.9$\pm$2493.3 & 666.5$\pm$354.5 & 591.7$\pm$533.4 & 823.2$\pm$344.4 \\
        MAPE & 2725$\pm$34378 & 10396$\pm$146605 & 11598$\pm$197033 & 6027$\pm$81126 & 12028$\pm$194671 \\
        RMSE & 874$\pm$727 & 1370$\pm$3288 & 1160$\pm$607 & 960$\pm$781 & 1144$\pm$572 \\
        P.Coef. & 0.121$\pm$0.094 & 0.061$\pm$0.048 & 0.057$\pm$0.045 & 0.090$\pm$0.074 & 0.052$\pm$0.085 \\
        SSIM & 0.349$\pm$0.117 & 0.246$\pm$0.096 & 0.232$\pm$0.097 & 0.299$\pm$0.110 & 0.232$\pm$0.069 \\
        \hline
    \end{tabular}
    \caption{Error metrics of the various tested Models: Mean Absolute Error (MAE), Mean Absolute Percentage Error (MAPE), Root Mean Squared Error (RMSE), Pearson Correlation (P.Coef), Structural similarity index (SSIM)}
    \label{tab:error_stats}
\end{table}

\begin{table}[ht]
    \centering
    \begin{tabular}{cccccc}
        \hline
        Model& $\sum_{Eukl. dist.}$ & Avg. & min. & max. & Median \\
        \hline
        WGAN & 3430.962 & 1.9008 & 0.0089 & 8.2103  & 1.4706 \\
        DDPM & 13519.116 & 7.4898 & 0.0019 & 25.3349 & 3.1912 \\
        HMM  & 7334.314 & 4.0633 & 0.0096 & 21.3038 & 3.3176 \\
        MABF  & 3882.653 & 2.1511 & 0.0067 & 11.1387 & 1.2211 \\
        SLP  & 16157.287 & 8.9514 & 0.0495 & 24.7531 & 8.9684 \\
    \hline
    \end{tabular}
    \caption{UMAP-Analysis: Statistics of Euklidean distances of Hungarian matched daily power consumption samples. Summed pairwise distance, minimum, maximum and median distance.}
    \label{tab:umap_table}
\end{table}

\begin{figure}
    \centering
    \includegraphics[width=0.95\linewidth]{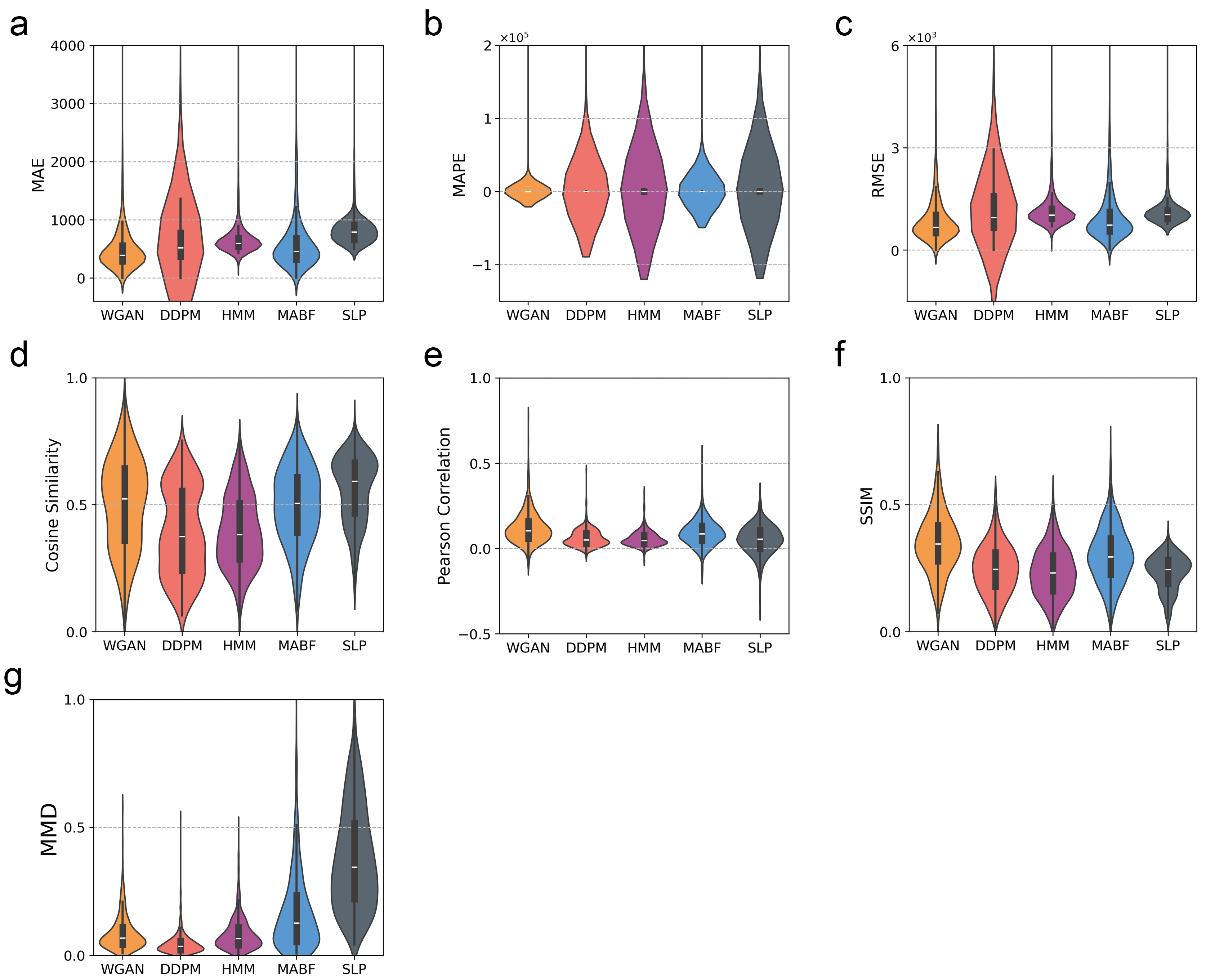}
    \caption{Error metrics - Violin-plots for error metrics of the various tested Models: a)Mean Absolute Error (MAE), b) Mean Absolute Percentage Error (MAPE), c) Root Mean Squared Error (RMSE), d) Cosine Similarity (Cos SI), e) Pearson Correlation (P.Coef), f) Structural similarity index (SSIM) and g) Maximum mean discrepancy (MMD) b)~Quantitative comparison of real and synthetic samples using Maximum Mean Discrepancy (MMD) to evaluate distributional similarity. Lower MMD values indicate higher fidelity. \gls{DDPM} and \gls{WGAN} achieved the highest average similarity. \gls{MABF} showed a markedly reduced similarity, whereas the \gls{SLP} samples had the lowest resemblence to the real distribution.}
    \label{fig:error-metrics}
\end{figure}

\begin{figure}[!ht]
\centering
\resizebox{0.85\textwidth}{!}{\includegraphics{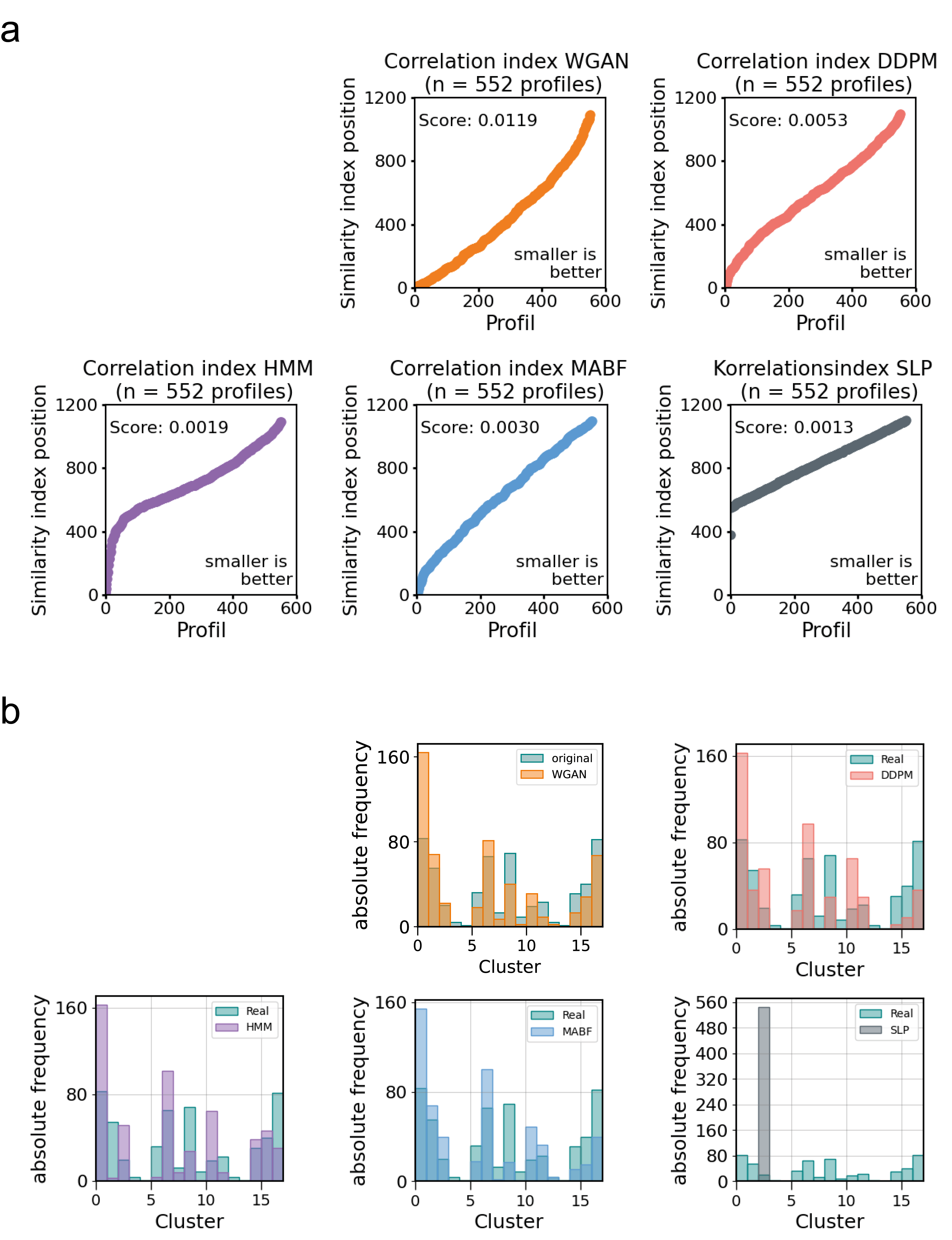}}
\caption{Similarity analysis - Correlation index and matching frequency. a) Correlation index (smaller is better, ideally 1) to assess the similarity of a digital surrogate to its native real-world consumer. All profiles were analyzed and an overall Score provided for each tested ML model (0 - 1, higher is better). b) Absolute frequency of the most abundant consumer type found in the ten most similar consumers from the similarity index position (in a) ) for each ML model.}
\label{fig:corr_analysis2}
\end{figure}

\begin{figure}
    \centering
    \includegraphics[width=0.85\linewidth]{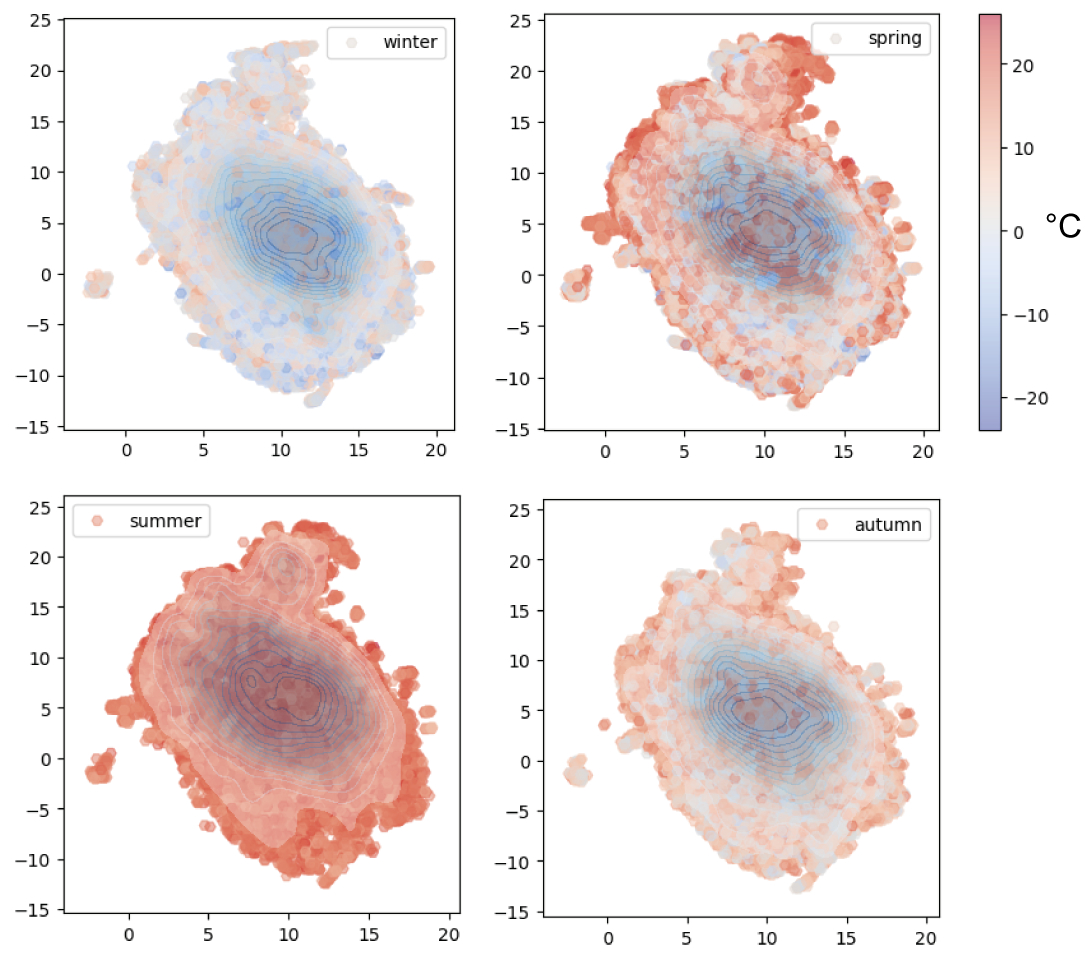}
    \caption{UMAP analysis of single day samples from the \gls{om} dataset with average day-temperature mapping for the respective sample and a KDE-plot on top to visualize density. No clear clusters for specific temperature ranges or seasons can be detected.}
    \label{fig:umap-temperature}
\end{figure}

\end{document}